\documentclass[10pt,journal]{IEEEtran}

\usepackage{amsmath,amssymb,mathtools,amsthm}
\usepackage{graphicx}
\usepackage{booktabs}
\usepackage{multirow}
\usepackage{cite}
\usepackage{xcolor}
\usepackage{url,bm}

\newtheorem{theorem}{Theorem}
\newtheorem{proposition}{Proposition}
\newtheorem{corollary}{Corollary}

\newcounter{algorithm}
\newenvironment{algorithmblock}[1]{%
\begin{figure}[t]\refstepcounter{algorithm}\small
\hrule\vspace{2pt}\noindent\textbf{Algorithm \thealgorithm: #1}\vspace{2pt}\hrule\vspace{2pt}}
{\vspace{2pt}\hrule\end{figure}}

\newcommand{\Tr}{\operatorname{tr}}
\newcommand{\rank}{\operatorname{rank}}
\newcommand{\diag}{\operatorname{diag}}
\newcommand{\supp}{\operatorname{supp}}
\newcommand{\Cov}{\operatorname{Cov}}

\newcommand{\cN}{\mathcal{N}}
\newcommand{\cU}{\mathcal{U}}

\newcommand{\eps}{\epsilon}
\newcommand{\e}{\mathrm{e}}
\newcommand{\dd}{\mathrm{d}}
\newcommand{\trans}{\mathsf{T}}

\newcommand{\simfigure}[4]{%
\begin{figure}[t]
  \centering
  \IfFileExists{#1}{\includegraphics[width=\columnwidth]{#1}}{%
  \fbox{\parbox[c][1.45in][c]{0.94\columnwidth}{\centering\small #4}}}
  \caption{#2}
  \label{#3}
\end{figure}}

\newcommand{\simfigurewide}[4]{%
\begin{figure*}[t]
  \centering
  \IfFileExists{#1}{\includegraphics[width=0.96\textwidth]{#1}}{%
  \fbox{\parbox[c][1.65in][c]{0.92\textwidth}{\centering\small #4}}}
  \caption{#2}
  \label{#3}
\end{figure*}}

\RequirePackage{graphicx}
\makeatletter
\@ifpackageloaded{subfig}{%
  \PackageError{MIC figures}{Do not load subfig together with subfigure}%
  {Remove the subfig package before loading figure_setup.tex.}%
}{}
\@ifpackageloaded{subcaption}{%
  \PackageError{MIC figures}{Do not load subcaption together with subfigure}%
  {Remove the subcaption package before loading figure_setup.tex.}%
}{}
\@ifpackageloaded{subfigure}{}{\RequirePackage[tight,footnotesize]{subfigure}}
\makeatother
\begin{document}

\title{Mobility Information Capacity of Multi-Robot Systems: Safety-Aware Maneuver Design}

\author{Weijie Yuan,~\IEEEmembership{Senior Member,~IEEE}
\thanks{The author is with the School of Automation and Intelligent Manufacturing, Southern University of Science and Technology, Shenzhen, China (yuanwj@sustech.edu.cn).}}

\maketitle

\begin{abstract}
Multiple robots operating in shared workspaces must adjust their speed, direction, and trajectories to coordinate with one another and respond to changing tasks and disturbances. The available maneuver freedom depends on the dynamics, uncertainty, and safety margins, even for a fixed fleet size and workspace. However, robot count, workspace occupancy, and task throughput do not quantify how much the fleet can vary its motion or how reliably these variations can be observed. We develop a mobility information capacity (MIC) framework that jointly characterizes these two aspects. Mobility information is defined as the mutual information between a joint maneuver sequence and its noisy trajectory observation, and MIC maximizes this quantity under maneuver-resource constraints. For a linear Gaussian maneuver-to-observation model with feedback dynamics, we characterize the finite-horizon information benchmark and derive its stationary long-horizon spectral limit, accounting for heterogeneous mobility gains, colored uncertainty, and temporally coupled maneuver costs. We further map workspace-boundary, inter-robot separation, and actuator chance constraints into coupled maneuver-covariance restrictions and derive conditions under which covariance optimization reduces to resource allocation across mobility modes. Finite joint maneuver codebooks are then constructed through individual risk screening and noise-whitened trajectory separation, with bounds linking sequence recovery error and achieved information to the capacity benchmarks. An aerial-robot simulation study evaluates spectral convergence, safety-aware allocation, finite-codebook identification, and fleet-size scaling. The resulting framework complements occupancy and task-throughput measures by quantifying the maximum observable maneuver freedom supported by a mobility system.

\end{abstract}

\begin{IEEEkeywords}
Multi-robot system, mobile computing, aerial robot, information theory, chance constraints.
\end{IEEEkeywords}

\section{Introduction}
\begin{figure*}[t]
    \centering
    \includegraphics[width=0.90\textwidth]{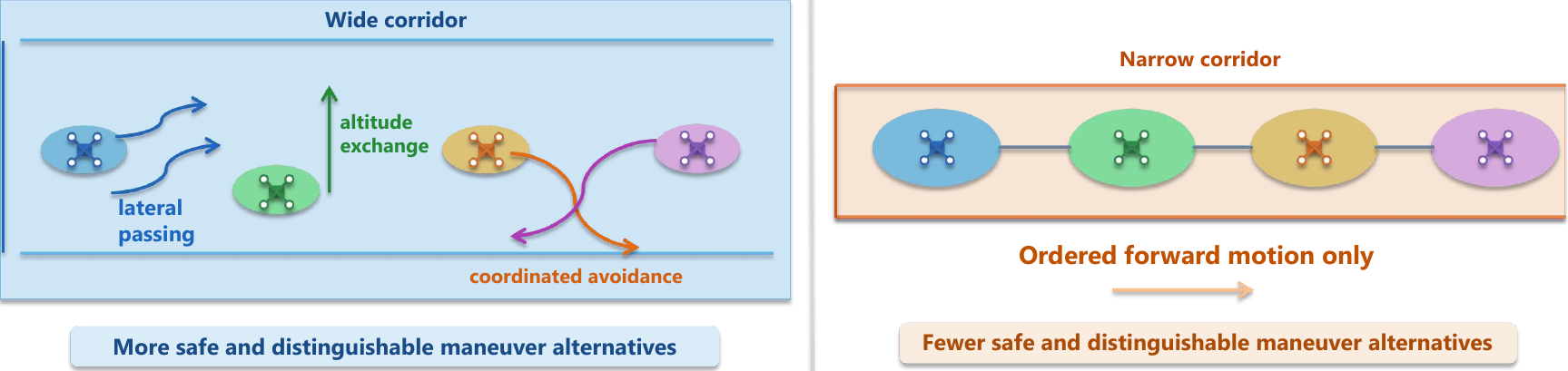}
    \caption{Same occupancy can lead to substantially different operational mobility.}
    \label{fig:airspace_capacity_concept}
\end{figure*}
Multi-robot systems increasingly operate in shared, spatially constrained environments, where changing tasks, disturbances, and neighboring robots require continual motion adjustments \cite{garg2024safe}. Even with the same workspace and fleet size, systems with different dynamics, uncertainty, and safety margins may support different sets of safe maneuvers, including acceleration, deceleration, direction changes, and rerouting. Robot count measures occupancy, while task throughput measures completed work. Although both are important for task planning, neither fully characterizes the maneuver flexibility available to the fleet. Such a characterization must account for both physical maneuverability and trajectory observability: a constrained workspace may allow little safe maneuver variation despite accurate observations, while observation uncertainty may make otherwise feasible maneuvers difficult to distinguish.

Aerial-robot fleets illustrate this distinction. Unmanned aerial vehicles (UAVs) combine mobility, sensing, communication, and control in low-altitude applications \cite{motlagh2016,mozaffari2017,meng2023,jin2025tmc,wu2024isac}. As shown in Fig.~\ref{fig:airspace_capacity_concept}, two corridors can accommodate the same number of aerial robots while permitting different lateral and vertical maneuvers. A wide corridor may allow lateral passing, altitude exchange, and several coordinated avoidance patterns, whereas a narrow corridor may permit only ordered forward motion. The information conveyed by these maneuvers also depends on the accuracy of the trajectory observations. Physical constraints determine the maneuver variation that can be generated, while observation uncertainty determines how much of that variation remains distinguishable. This leads to the following question: how much mobility information can a multi-robot system convey through its trajectories under prescribed maneuver-resource and safety constraints?

To address this question, we model the joint maneuver sequence as the input and the observed fleet trajectory as the output, with the feedback architecture fixed across maneuver alternatives. Execution disturbances affect the physical motion, while observation errors affect the trajectories available to the mobility monitor. For a prescribed maneuver distribution, mutual information quantifies how much the observed trajectory reveals about the underlying maneuver sequence. MIC maximizes this quantity over admissible maneuver distributions under a resource budget. It therefore characterizes the maximum observable maneuver information supported by the specified dynamics, uncertainty, and maneuver constraints.



\subsection{Relation to Existing Metrics}
MIC is related to existing information-theoretic and control-theoretic measures through its use of mutual information, but differs in its operational interpretation and system construction. Shannon capacity quantifies the reliably distinguishable alternatives supported by a channel under a resource constraint \cite{shannon1948,cover2006}. Action-to-state mutual information has also been used to characterize control-to-state capacity \cite{klyubin2005,tiomkin2024,levy2025representation}. A recent multi-agent formulation models coupled actions as an interference channel and applies iterative water-filling to study collective behavior \cite{shah2026multiagent}. MIC adopts this information-theoretic perspective to quantify the distinguishability of joint maneuvers supported by a specified mobility architecture. Its objective differs from that of stabilization-oriented information measures, which quantify the information required to maintain a control objective \cite{nair2004}.

The aerial-robot application further connects MIC to studies of airspace capacity and mobile computing. Urban airspace studies address traffic assignment and route utilization, while UAV mobile computing studies optimize trajectories, communication, computation, and service resources \cite{wang2021airtraffic,sun2024multiuavmec,wei2026airfogsim}. MIC complements these application-level measures by quantifying the maneuver information recoverable from trajectory observations. For aerial workspaces, this provides a maneuver-information characterization of airspace use. More generally, MIC depends on robot dynamics, observation uncertainty, maneuver cost, and physical safety constraints.

\subsection{Contributions}

This paper extends the finite-horizon Gaussian MIC framework in our preliminary letter \cite{yuan2026mobility} to safety-aware multi-robot systems. We investigate how observable maneuver information grows with the horizon, how physical constraints reshape maneuver allocation, and how continuous covariance designs support finite codebooks with individually screened maneuvers. The main contributions are summarized as follows.

\begin{itemize}

\item We derive a heterogeneous stationary spectral characterization of swarm MIC under stable closed-loop dynamics, colored uncertainty, and temporally coupled maneuver costs. This characterization identifies the maneuver components that sustain a nonzero information rate and determines resource allocation across frequencies and mobility modes.

\item We formulate safety-aware joint maneuver allocation by mapping workspace-boundary, inter-robot separation, and actuator chance constraints into coupled covariance restrictions. The design jointly accounts for trajectory observability and physical safety margins. We derive sufficient conditions for optimal allocation in a common mobility-mode basis and the corresponding water-filling rule. For general constraints, the formulation also optimizes correlations between mobility modes.

\item We construct finite joint maneuver codebooks through individual risk screening and noise-whitened trajectory packing. Every retained sequence satisfies the prescribed individual risk and deterministic maneuver-cost limits. We derive an identification-error bound and relate the achieved mutual information to the individually safe capacity and the unconstrained Gaussian benchmark.
\end{itemize}

An aerial-robot case study evaluates the predicted spectral convergence, the safety--information tradeoff, and the codeword error performance of individually risk-screened codebooks. It also examines how fleet-wide and per-robot information vary with fleet size.

The remainder of this paper is organized as follows. Section~\ref{sec:model} presents the operational system model. Sections~\ref{sec:finite} and \ref{sec:spectral} derive the finite- and long-horizon capacity laws, respectively. Section~\ref{sec:safety} develops safe capacity, covariance allocation, and finite codebook construction. Section~\ref{sec:simulation} presents the aerial-robot case study, and Section~\ref{sec:discussion} concludes the paper.

\emph{Notation:} Bold lowercase and uppercase letters denote vectors and matrices, respectively. For a real symmetric matrix $\mathbf X$, $\mathbf X\succeq\mathbf0$ denotes positive semidefiniteness. The operators $(\cdot)^{\trans}$, $\Tr(\cdot)$, and $\rank(\cdot)$ denote transpose, trace, and rank, respectively. The superscript $(\cdot)^*$ denotes conjugate transpose, $\|\cdot\|_2$ denotes the Euclidean norm for vectors and the spectral norm for matrices, and $\|\cdot\|_{\mathrm F}$ denotes the Frobenius norm. The functions $\Phi(\cdot)$ and $Q(\cdot)=1-\Phi(\cdot)$ denote the standard-normal cumulative distribution function and upper-tail function, respectively. All logarithms in information quantities are to base two.

\section{Multi-robot System and Operational Model}
\label{sec:model}

\subsection{Coupled Fleet Trajectory Observation}

\begin{figure}[t]
    \centering
    \includegraphics[width=0.50\textwidth]{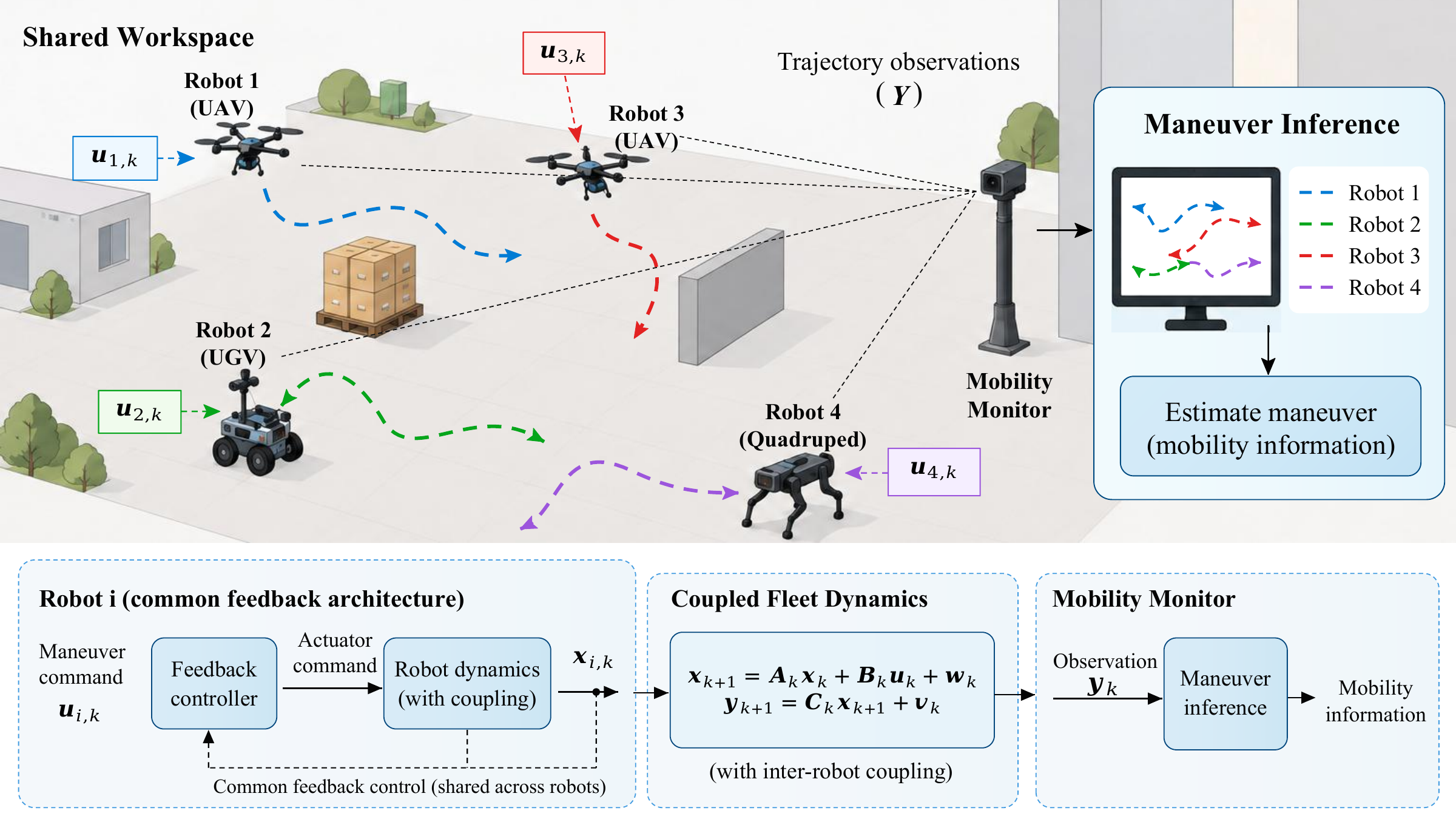}
    \caption{Multi-robot operational model.}
    \label{fig:model}
\end{figure}

We consider $N$ mobile robots operating in a shared workspace over $K$ time steps of duration $\Delta t$, with horizon $T=K\Delta t$, as illustrated in Fig.~\ref{fig:model}. The state of robot $i$ may include planar (2D) or spatial (3D) position, velocity, orientation, and controller variables. We stack the fleet states and intentional maneuver commands at time $k$ as $\mathbf{x}_k=[\mathbf{x}_{1,k}^{\trans},\ldots,\mathbf{x}_{N,k}^{\trans}]^{\trans}$ and $\mathbf{u}_k=[\mathbf{u}_{1,k}^{\trans},\ldots,\mathbf{u}_{N,k}^{\trans}]^{\trans}$, where $m_u=\dim(\mathbf{u}_k)$. Each robot operates under the shared feedback architecture, while a mobility monitor observes the robots' noisy trajectories to obtain information about the underlying joint maneuver sequence. We assume that the monitor knows the maneuver-to-observation model. Denoting the trajectory observation by $\mathbf{y}_k$ at time $k$, we write the joint linear model as
\begin{align}
\mathbf{x}_{k+1}&=\mathbf{A}_k\mathbf{x}_k+\mathbf{B}_k\mathbf{u}_k+\mathbf{w}_k, \label{eq:state}\\
\mathbf{y}_k&=\mathbf{C}_k\mathbf{x}_{k+1}+\mathbf{v}_k, \label{eq:observation}
\end{align}
The diagonal blocks of $\mathbf{A}_k$ describe the closed-loop responses of individual robots, whereas its off-diagonal blocks capture coupling through the dynamics, formation control, or coordination. The matrix $\mathbf B_k$ may also couple maneuver inputs across robots, and the process disturbances may be correlated across the swarm. Together, these matrices specify the motion capabilities of the selected robot platforms, including their fixed feedback dynamics. For example, if the physical actuator command is $\mathbf{a}_k=\mathbf{K}_k\mathbf{x}_k+\mathbf{E}_k\mathbf{u}_k$, substituting it into the plant model yields \eqref{eq:state}. Thus, $\mathbf{u}_k$ is a joint, mission-dependent mobility input applied outside the common feedback law; it is not a causal feedback policy adapted to past disturbances or observations. The vector $\mathbf y_k$ denotes the noisy trajectory observation available to the mobility monitor, and $\mathbf v_k$ denotes the additive observation error.

Define the transition matrix
\begin{equation}
\mathbf{T}(k,j)=
\begin{cases}
\mathbf{A}_{k-1}\mathbf{A}_{k-2}\cdots\mathbf{A}_{j}, & k>j,\\
\mathbf{I}, & k=j.
\end{cases}
\end{equation}
Stacking $\mathbf{U}=[\mathbf{u}_0^{\trans},\ldots,\mathbf{u}_{K-1}^{\trans}]^{\trans}$ and $\mathbf{Y}=[\mathbf{y}_1^{\trans},\ldots,\mathbf{y}_K^{\trans}]^{\trans}$ yields the lifted model
\begin{equation}
\mathbf{Y}=\mathbf{F}_K\mathbf{x}_0+\mathbf{G}_K\mathbf{U}+\mathbf{Z}_K,
\label{eq:lifted}
\end{equation}
In \eqref{eq:lifted}, $\mathbf{Z}_K\sim\cN(\mathbf{0},\mathbf{K}_{Z,K})$, $\mathbf{F}_K$ denotes the lifted initial-state response, and $\mathbf{G}_K$ is the maneuver-to-observation matrix. Its $(r,s)$ block is given by
\begin{equation}
[\mathbf{G}_K]_{r,s}=
\begin{cases}
\mathbf{C}_r\mathbf{T}(r,s+1)\mathbf{B}_s, & s\le r-1,\\
\mathbf{0}, & s\ge r.
\end{cases}
\label{eq:Gblocks}
\end{equation}
Since the initial state $\mathbf{x}_0$ is known, the first term on the right-hand side of \eqref{eq:lifted} can be subtracted. The effective uncertainty $\mathbf{Z}_K$ comprises propagated process disturbances and observation error, which may be correlated across robots and time. Throughout the Gaussian analysis, we assume that $\mathbf Z_K$ is independent of $\mathbf U$ and has positive-definite covariance $\mathbf K_{Z,K}$.

\subsection{Maneuver Resource and Operational Interpretation}
To establish an operational coding interpretation, we introduce the index $J\in \{1,...,M\}$. When \(J=m\), the exogenous maneuver input is \(\mathbf u^{(m)}\). From the observation \(\mathbf Y\), the monitor identifies the selected sequence and estimates the corresponding index as \(\widehat J\), with identification error
\begin{equation}
P_{\mathrm e}=\Pr(\widehat{J}\ne J)\le\eta.
\label{eq:error}
\end{equation}
All maneuver alternatives use the same controller and differ only in their exogenous input sequences. Each physical trajectory is a random realization induced by $\mathbf{u}^{(m)}$ and the disturbance.

We use $\mathbf{Q}_K\succ\mathbf{0}$ as the maneuver-cost matrix and impose the resource constraint
\begin{equation}
\mathbb{E}[\mathbf{U}^{\trans}\mathbf{Q}_K\mathbf{U}]\le P_M T.
\label{eq:budget}
\end{equation}
The matrix $\mathbf Q_K$ weights intentional maneuver effort, such as acceleration variation or deviation from a nominal command sequence, and may penalize rapid changes between time steps. The scalar $P_M$ denotes the mobility-resource rate, and $P_MT$ is the total resource budget over horizon $T$.

\subsection{Safety Semantics}
For a given maneuver sequence $\mathbf{u}$, we assess safety through a collection of violation events. We consider three types of violations: a robot leaving its admissible region, a pair of robots approaching closer than $d_{\min}$, and a violation of prescribed actuator or other operational limits. Let $\mathcal E_r(\mathbf u)$ denote the event that the $r$-th constraint is violated. The union of these events is
\begin{equation}
\mathcal E_{K}(\mathbf u)
=\bigcup_{r=1}^{R}\mathcal E_r(\mathbf u).
\label{eq:mic_sampled_event}
\end{equation}

For a given risk tolerance $\epsilon$, we define the safe maneuver set as
\begin{equation}
\mathcal{U}_{\epsilon}
=
\left\{\mathbf{u}:\Pr\!\left[\mathcal{E}_K(\mathbf{u})\right]
\le \epsilon
\right\},
\label{eq:safeset}
\end{equation}
where the probability is taken over the execution uncertainty for a fixed maneuver $\mathbf{u}$. Thus, $\epsilon$ characterizes the physical execution risk of an individual maneuver, and $\mathcal{U}_{\epsilon}$ specifies the desired horizon-wise safety criterion. Section~\ref{sec:safety} develops a tractable chance-constrained formulation that incorporates this requirement into MIC optimization.

\section{Fleet-Level Finite-Horizon Benchmark}
\label{sec:finite}

\subsection{Coupled Fleet Log-Determinant}
We first extend the finite-horizon Gaussian benchmark in our preliminary letter to the multi-robot system. The resulting information objective underlies the subsequent spectral analysis and safety-aware design. For the centered channel $\mathbf{Y}=\mathbf{G}_K\mathbf{U}+\mathbf{Z}_K$, we define the uncertainty-whitened maneuver-to-trajectory mapping $\mathbf{H}_K=\mathbf{K}_{Z,K}^{-1/2}\mathbf{G}_K$. For any admissible maneuver distribution \(p(\mathbf U)\), the mobility information $I(\mathbf U;\mathbf Y)$ over the \(T\)-horizon quantifies the information that the observed trajectory carries about the underlying maneuver sequence. We define MIC as the maximum mobility information under the resource constraint, i.e.,
\begin{equation}
C_M(T)=\sup_{p(\mathbf{U}):\,\mathbb{E}[\mathbf{U}^{\trans}\mathbf{Q}_K\mathbf{U}]\le P_MT} I(\mathbf{U};\mathbf{Y}).
\label{eq:capacitydef}
\end{equation}
Thus, $C_M(T)$ is measured in bits per realization over horizon $T$. Relative to the differential entropy $h(\mathbf{Y})$, mutual information removes the uncertainty that remains when the intentional input is known, i.e.,
\begin{equation}
I(\mathbf{U};\mathbf{Y})=h(\mathbf{Y})-h(\mathbf{Y}\mid\mathbf{U})=h(\mathbf{Y})-h(\mathbf{Z}_K).
\end{equation}
Accurate observation alone therefore does not imply high mobility information when the admissible maneuver distribution contains little variation.

\begin{proposition}[Finite-horizon Gaussian fleet benchmark]
\label{prop:finite}
For the model in \eqref{eq:lifted} under the constraint in \eqref{eq:budget}, an optimal input is zero-mean Gaussian. The maneuver-input covariance is $\mathbf{K}_U=\Cov(\mathbf{U})$. Under the Gaussian channel model, $C_M(T)=C_M^{\mathrm G}(T)$, where
\begin{align}
C_M^{\mathrm G}(T)=\max_{\substack{\mathbf{K}_U\succeq\mathbf{0}:\\\Tr(\mathbf{Q}_K\mathbf{K}_U)\le P_MT}}
\frac{1}{2}\log_2\det\!\left(\mathbf{I}+\mathbf{H}_K\mathbf{K}_U\mathbf{H}_K^{\trans}\right).
\label{eq:logdet}
\end{align}
\end{proposition}

\begin{IEEEproof}
The Gaussian-extremality argument extends \cite[Prop.~1]{yuan2026mobility} to the fleet-matrix setting. A nonzero input mean consumes joint maneuver resources but only translates the output; hence, the optimal mean is zero. For fixed $\mathbf{K}_U$, the Gaussian maximum-entropy property provides an upper bound on $h(\mathbf{Y})$, which is attained by a Gaussian fleet input. Independence across robots is not assumed, and the cross-robot covariance entries of $\mathbf{K}_U$ remain optimization variables.
\end{IEEEproof}

The determinant in \eqref{eq:logdet} measures the extent to which admissible maneuver variation expands the uncertainty-whitened trajectory space. A high value requires both rich maneuver variation and distinguishable trajectory responses. This benchmark accounts for maneuver resources; physical safety constraints are incorporated in Section~\ref{sec:safety}.

\subsection{Finite-Horizon Mobility Modes}

To characterize the joint maneuver directions, we define the cost- and uncertainty-whitened channel
\begin{equation}
\widetilde{\mathbf{G}}_K=\mathbf{K}_{Z,K}^{-1/2}\mathbf{G}_K\mathbf{Q}_K^{-1/2}.
\label{eq:whitened}
\end{equation}
Let $\mathbf v_\ell$ denote a unit eigenvector of $\widetilde{\mathbf G}_K^{\trans}\widetilde{\mathbf G}_K$ associated with $\lambda_\ell>0$, where $L_K=\rank(\widetilde{\mathbf{G}}_K)$. We refer to $\mathbf v_\ell$ as a cost-normalized mobility-mode vector specifying a normalized spatiotemporal maneuver pattern, and to $\lambda_\ell$ as its mobility gain. Applying the singular-value decomposition of $\widetilde{\mathbf{G}}_K$ and Hadamard's inequality diagonalizes the optimal covariance and gives
\begin{equation}
C_M^{\mathrm G}(T)=
\max_{\substack{p_\ell\ge0:\,\sum_{\ell=1}^{L_K}p_\ell\le P_MT}}
\frac{1}{2}\sum_{\ell=1}^{L_K}\log_2(1+\lambda_\ell p_\ell),
\label{eq:parallel}
\end{equation}
with
\begin{equation}
p_\ell^\star=\left(\nu-\frac{1}{\lambda_\ell}\right)^+,
\qquad \sum_{\ell=1}^{L_K}p_\ell^\star=P_MT.
\label{eq:wf}
\end{equation}

For the whitened channel $\widetilde{\mathbf{G}}_K$, each right singular vector $\mathbf{v}_\ell$ has an associated mobility gain
\begin{equation}
\lambda_\ell=\|\widetilde{\mathbf{G}}_K\mathbf{v}_\ell\|_2^2
\end{equation}
that measures the observable trajectory variation produced by one unit of normalized maneuver input relative to uncertainty. A mode can combine the motion of several robots with coordinated corrections over multiple time steps. It specifies a direction of joint maneuver variation rather than a selected mission or physical trajectory. The allocation in \eqref{eq:wf} assigns maneuver resources to patterns that the fleet can produce economically and the mobility monitor can distinguish reliably. The mobility modes therefore identify maneuver directions that yield the largest observable motion variation.


\subsection{Operational Interpretation of the Finite-Horizon Benchmark}

The finite-horizon quantity $C_M^{\mathrm G}(T)$ measures the mutual information between a complete joint maneuver over $[0,T]$ and its trajectory observation. To connect this quantity to reliable identification, we treat each $T$-horizon maneuver-observation pair as one use of a vector channel. For $n$ independent uses, we have
\begin{equation}
\mathbf{Y}_b=\mathbf{G}_K\mathbf{U}_b+\mathbf{Z}_b, \qquad b=1,\ldots,n,
\label{eq:blocks}
\end{equation}
where $\mathbf{Z}_b$ are independent copies of $\mathbf{Z}_K$. An $(n,M,\eta)$ mobility code uses a random label $J$. For a realization $J=m$, the corresponding codeword consists of a deterministic sequence of $n$ joint fleet maneuvers $\{\mathbf{u}_b^{(m)}\}_{b=1}^{n}$, with $\mathbf{U}_b=\mathbf{u}_b^{(m)}$ for $b=1,\ldots,n$. The mobility monitor jointly estimates $J$ from $\mathbf{Y}^n=(\mathbf{Y}_1,\ldots,\mathbf{Y}_n)$. The rate $R=n^{-1}\log_2 M$ measures the logarithmic number of reliably distinguishable joint maneuver codewords per \(T\)-horizon block.

For $0<\eta<1/2$, the code has average identification error $\Pr(\widehat{J}\ne J)\le\eta$ and satisfies
\begin{equation}
\frac{1}{Mn}
\sum_{m=1}^{M}\sum_{b=1}^{n}
(\mathbf{u}_b^{(m)})^{\trans}\mathbf{Q}_K\mathbf{u}_b^{(m)}
\le P_MT.
\label{eq:codecost}
\end{equation}
The constraint averages resource consumption over the uniformly selected codewords and the \(n\) maneuver blocks. Accordingly, the coding rate is $n^{-1}\log_2 M$ bits per $T$-horizon block. Let $M^\star(n,T,\eta)$ denote the largest code size satisfying these requirements.

\begin{theorem}[Repeated-block operational capacity]
\label{thm:operational}
For any fixed $T$, the mobility channel subject to \eqref{eq:codecost} satisfies
\begin{equation}
\lim_{\eta\to0^+}\limsup_{n\to\infty} \frac{1}{n}\log_2 M^\star(n,T,\eta) =C_M^{\mathrm{G}}(T).
\label{eq:coding}
\end{equation}
Thus, every rate below $C_M^{\mathrm{G}}(T)$ is achievable with vanishing average identification error, whereas no rate strictly above it is achievable with vanishing error.
\end{theorem}

\begin{IEEEproof}
Achievability follows from the standard random-coding theorem for a memoryless channel under an average cost constraint. For any $R<C_M^{\mathrm G}(T)$, we can choose a zero-mean Gaussian input with maneuver cost strictly below $P_MT$ and mutual information exceeding $R$. The usual expurgation argument establishes the existence of a deterministic codebook satisfying the empirical average-cost constraint. For the converse, let $J$ select an admissible codeword over $n$ independent horizon blocks. With $h_2(p)=-p\log_2p-(1-p)\log_2(1-p)$, Fano's inequality and the concavity of the Gaussian information objective give
\begin{align}
(1-\eta)&\log_2 M
\le I(J;\mathbf Y^n)+h_2(\eta)\nonumber\\
&\le \sum_{b=1}^{n} I(\mathbf U_b;\mathbf Y_b)+h_2(\eta)
\le nC_M^{\mathrm G}(T)+h_2(\eta).
\end{align}
The last inequality follows because the average input covariance satisfies the maneuver budget. Dividing by $n$ and taking $n\to\infty$ followed by $\eta\to0^+$ proves the result; see \cite[Ch.~7]{cover2006}.
\end{IEEEproof}

Theorem~\ref{thm:operational} establishes the operational meaning of the finite-horizon log-determinant: $C_M^{\mathrm G}(T)$ characterizes the asymptotic distinguishability per $T$-horizon maneuver-observation block. For any $R<C_M^{\mathrm G}(T)$, the number of reliably distinguishable joint maneuver codewords over $n$ independent horizon blocks can grow approximately as $2^{nR}$ while the average identification error vanishes as $n\rightarrow\infty$. The repetition index $n$ counts independent uses of the entire $N$-robot, $T$-horizon mobility channel, rather than robots. For a small finite number of independent maneuver episodes, this asymptotic result does not specify the number of labels achievable at a given identification error; that question requires a finite-blocklength (FBL) analysis. The next section examines whether mobility information grows proportionally with the physical horizon.

\section{Long-Horizon Mobility Capacity}
\label{sec:stationary}
\label{sec:spectral}

Section~\ref{sec:finite} characterizes mobility information over a prescribed finite trajectory horizon. For any fixed $K$, this formulation is exact: each gain $\lambda_\ell$ corresponds to a complete spatiotemporal maneuver pattern over a horizon of $T$, and $C_M^{\mathrm G}(K\Delta t)$ quantifies the information carried by the corresponding trajectory segment. As $K$ increases, however, both the number and the structure of these modes change. The finite-horizon mobility-mode representation alone therefore does not establish whether mobility information continues to grow in proportion to physical time.

We therefore develop a long-horizon description of the same mobility information problem. Under a stationary operating regime, the growing set of finite-horizon modes admits a frequency-domain characterization independent of the chosen horizon length. This representation determines whether $\frac{C_M^{\mathrm G}(K\Delta t)}{K}$ approaches a nonzero limit and, consequently, whether the system sustains a positive mobility information rate as the observation horizon grows.

\subsection{From Finite-Horizon to Stationary Mobility Modes}

To characterize the mobility modes independently of the horizon, we consider a stationary operating regime, such as steady reference tracking or formation motion. We set $\mathbf A_k=\mathbf A$, $\mathbf B_k=\mathbf B$, and $\mathbf C_k=\mathbf C$, and assume that all eigenvalues of $\mathbf A$ have modulus less than one, i.e., the dynamics are Schur-stable. The noise-free maneuver-to-observation map is then a stable causal convolution with transfer matrix
\begin{equation}
\mathbf{H}(\e^{\mathrm{j}\omega})
=
\mathbf{C}
\big(\e^{\mathrm{j}\omega}\mathbf{I}-\mathbf{A}\big)^{-1}
\mathbf{B}.
\label{eq:transfer}
\end{equation}
Thus, the lifted matrix $\mathbf{G}_K$ in Section~\ref{sec:model} and $\mathbf{H}(\e^{\mathrm{j}\omega})$ describe the same closed-loop mobility channel over a finite segment and in the frequency domain, respectively.

Let $\{\mathbf z_k\}$ be a zero-mean stationary Gaussian output-uncertainty process with power spectral density matrix $\mathbf S_Z(e^{\mathrm j\omega})$, where $\omega$ is measured in radians per time step. Its autocovariance matrices are
\begin{align}
\mathbf R_Z[l]
=\mathbb E[\mathbf z_{k+l}\mathbf z_k^T]
=\frac{1}{2\pi}\int_{-\pi}^{\pi}
 \mathbf S_Z(e^{j\omega})e^{j\omega l}\,d\omega.
\end{align}
Here, $\omega$ denotes the angular frequency in radians per time step, and $l$ is an integer time lag.
When the stationary process disturbance and observation noise are independent,
\begin{equation}
\mathbf{S}_Z(\e^{\mathrm{j}\omega})
=
\mathbf{H}_w(\e^{\mathrm{j}\omega})
\mathbf{S}_w(\e^{\mathrm{j}\omega})
\mathbf{H}_w(\e^{\mathrm{j}\omega})^{*}
+
\mathbf{S}_v(\e^{\mathrm{j}\omega}),
\label{eq:noise_psd}
\end{equation}
where $\mathbf{H}_w(\e^{\mathrm{j}\omega})=\mathbf{C}(\e^{\mathrm{j}\omega}\mathbf{I}-\mathbf{A})^{-1}$. The matrices $\mathbf{S}_w(\e^{\mathrm{j}\omega})$ and $\mathbf{S}_v(\e^{\mathrm{j}\omega})$ denote the power spectral densities of $\mathbf{w}_k$ and $\mathbf{v}_k$, respectively. Equation~\eqref{eq:noise_psd} separates uncertainty propagated through the dynamics from observation uncertainty introduced after the dynamics.

Similarly, we assume that the quadratic maneuver cost depends only on the time separation between inputs. Let $\mathbf Q_l$ characterize the cross term between maneuver inputs separated by $l$ time instants. Nonzero $\mathbf Q_l$ for $l\neq 0$ allows the cost to penalize temporal variations in the maneuver sequence, such as rapid changes between consecutive acceleration commands. The finite-horizon cost matrix then takes the block-Toeplitz form \cite{bolten2022toeplitz}
\begin{equation}
\mathbf Q_K
=
\big[\mathbf Q_{r-s}\big]_{r,s=0}^{K-1}.
\end{equation}
Equivalently, the frequency-domain matrix $\mathbf Q(e^{\mathrm{j}\omega})$ specifies the maneuver cost associated with each frequency component, with
\begin{equation}
\mathbf Q_l
=
\frac{1}{2\pi}
\int_{-\pi}^{\pi}
\mathbf Q(e^{\mathrm{j}\omega})
e^{\mathrm{j}\omega l}\,
\mathrm d\omega,
\qquad
\mathbf Q_{-l}=\mathbf Q_l^{\trans}.
\label{eq:cost_spectrum}
\end{equation}
If the cost contains no cross-time terms, $\mathbf Q(e^{\mathrm{j}\omega})=\mathbf Q_0$ for all $\omega$.
Penalties on rapid maneuver variations, however, generally lead to a frequency-dependent $\mathbf Q(e^{\mathrm{j}\omega})$. We define the maneuver-resource budget per time instant as $P_s=P_M\Delta t$. Thus, $KP_s=P_MT$, consistent with the finite-horizon budget in \eqref{eq:budget}.

Normalizing by both observation uncertainty and maneuver cost gives the frequency-domain mobility matrix
\begin{align}
\mathbf{L}(\e^{\mathrm{j}\omega})
={}&
\mathbf{Q}(\e^{\mathrm{j}\omega})^{-1/2}
\mathbf{H}(\e^{\mathrm{j}\omega})^{*}
\mathbf{S}_Z(\e^{\mathrm{j}\omega})^{-1}
\nonumber\\
&\times
\mathbf{H}(\e^{\mathrm{j}\omega})
\mathbf{Q}(\e^{\mathrm{j}\omega})^{-1/2}.
\label{eq:symbol}
\end{align}
Let $\lambda_1(\omega),\ldots,\lambda_{m_u}(\omega)\geq0$ denote the eigenvalues of $\mathbf L(\e^{\mathrm{j}\omega})$. These spectral gains characterize the same cost- and uncertainty-normalized maneuver observability as the finite-horizon gains in Section~\ref{sec:finite}, but at a different time scale. A finite-horizon gain $\lambda_\ell$ corresponds to a complete spatiotemporal maneuver pattern over $K$ time instants and therefore depends on the chosen horizon. By contrast, $\lambda_\iota(\omega)$ characterizes a recurring maneuver component at temporal frequency $\omega$ in the stationary regime, independently of any particular $K$. 

\subsection{Long-Horizon Capacity Law}
We now examine whether the finite-horizon Gaussian benchmark under the stationary model grows linearly with the physical horizon. To this end, we characterize the limit of $C_M^{\mathrm G}(K\Delta t)/K$ as $K\rightarrow\infty$.
Theorem~\ref{thm:spectral} expresses this limit in terms of the spectral mobility gains.

\begin{theorem}[Long-horizon stationary mobility information capacity]
\label{thm:spectral}
Assume that $\mathbf{S}_Z$ and $\mathbf{Q}$ are continuous, uniformly positive definite Hermitian matrices, and that the finite normalized Gram matrices $\widetilde{\mathbf G}_K^{\trans}\widetilde{\mathbf G}_K$ satisfy the finite-section regularity condition stated in Appendix~A of the supplemental material. Then, for every $P_s\geq0$, the normalized finite-horizon capacity converges to
\begin{align}
C_\infty(P_s)
&\triangleq
\lim_{K\to\infty}
\frac{C_M^{\mathrm G}(K\Delta t)}{K}
\nonumber\\
&=
\max_{\{p_\iota(\omega)\}}
\frac{1}{4\pi}
\int_{-\pi}^{\pi}
\sum_{\iota=1}^{m_u}
\log_2\!\left(1+\lambda_\iota(\omega)p_\iota(\omega)\right)
\dd\omega,
\label{eq:spectralcap}
\end{align}
where $p_\iota(\omega)\geq0$ and
\begin{equation}
\frac{1}{2\pi}
\int_{-\pi}^{\pi}
\sum_{\iota=1}^{m_u}p_\iota(\omega)\dd\omega
\leq P_s.
\label{eq:spectralbudget}
\end{equation}
If $P_s>0$ and $\{(\omega,\iota):\lambda_\iota(\omega)>0\}$ has nonzero measure, the optimal allocation is
\begin{equation}
p_\iota^\star(\omega)
=
\left(\mu-\frac{1}{\lambda_\iota(\omega)}\right)^+,
\label{eq:spectralwf}
\end{equation}
with zero allocation wherever $\lambda_\iota(\omega)=0$ and $\mu$ chosen to satisfy \eqref{eq:spectralbudget} with equality. If all branches are zero almost everywhere, then $C_\infty(P_s)=0$. The corresponding mobility information rate is $R_M=C_\infty(P_s)/\Delta t$ bits/s.
\end{theorem}

\begin{IEEEproof}
See Appendix~A of the supplemental material. 
\end{IEEEproof}

Together with Theorem~\ref{thm:operational}, Theorem~\ref{thm:spectral} characterizes the long-horizon scaling of the operational baseline. For every fixed $K$, $C_M^{\mathrm G}(K\Delta t)$ denotes the reliable information per maneuver block under independent block repetition. Normalizing this quantity by $K$ yields a limit of $C_\infty(P_s)$, which characterizes the mobility information per time instant as the physical horizon increases.

For $P_s>0$, a positive long-horizon rate requires nonzero mobility gain on a set of frequencies with nonzero bandwidth. A longer trajectory can create $O(K)$ distinguishable modes only when the closed-loop dynamics and the observation architecture continue to reveal new temporal structure. Reducing the time-step duration $\Delta t$ alone does not guarantee additional mobility degrees of freedom, since the dynamics, uncertainty, and maneuver cost change simultaneously.

\subsection{System Implications}
The spectral law provides the following implications for long-horizon mobility monitoring.

\begin{corollary}[Flat-mode Shannon-like law]
\label{cor:flat}
Suppose that exactly $r$ spectral mobility gains are positive almost everywhere on $[-\pi,\pi]$ and share the constant value $\lambda$. With $B_M=r/(2\Delta t)$ and $P_s=P_M\Delta t$,
\begin{equation}
R_M
=
B_M\log_2(1+\mathrm{MNR}),
\qquad
\mathrm{MNR}
=
\frac{\lambda P_M}{2B_M}.
\label{eq:shannonlike}
\end{equation}
If the active-mode gains are nonflat and satisfy $0<\lambda_{\min}\leq\lambda_\iota(\omega)\leq\lambda_{\max}$, the rate is bounded below and above by \eqref{eq:shannonlike} with $\lambda$ replaced by $\lambda_{\min}$ and $\lambda_{\max}$, respectively.
\end{corollary}
\begin{IEEEproof}
Since the $r$ active modes have the same constant gain, symmetry and concavity yield the uniform allocation $p_\iota^\star(\omega)=P_s/r$ over the active modes. Substitution into \eqref{eq:spectralcap} gives \eqref{eq:shannonlike}. 
\end{IEEEproof}
Equation~\eqref{eq:shannonlike} is the flat-mode special case of the spectral law. Here, $2B_M$ denotes the number of effective real mobility modes per second, as determined by the dynamics and the trajectory-observation model.

\begin{corollary}[Terminal-only zero-rate condition]
\label{cor:terminal}
Suppose that the monitor observes only a fixed-dimensional terminal output. If the number of nonzero normalized modes satisfies $L_K\leq L_{\max}$ and their gains are uniformly bounded by $\lambda_{\max}$, then
\begin{equation}
C_M^{\mathrm G}(K\Delta t)
\leq
\frac{L_{\max}}{2}
\log_2\!\left(1+\frac{\lambda_{\max}KP_s}{L_{\max}}\right)
=O(\log K),
\label{eq:terminal_bound}
\end{equation}
and hence $C_M^{\mathrm G}(K\Delta t)/(K\Delta t)\to0$.
\end{corollary}
\begin{IEEEproof}
The finite-horizon representation and Jensen's inequality give
\begin{align}
C_M^{\mathrm G}(K\Delta t) &\leq \frac{L_K}{2} \log_2\!\left(1+\frac{\lambda_{\max}}{L_K}\sum_{\ell=1}^{L_K}p_\ell^\star\right) \nonumber\\&\leq\frac{L_{\max}}{2}\log_2\!\left(1+\frac{\lambda_{\max}KP_s}{L_{\max}}\right).
\end{align}
Since $L_{\max}$ is independent of $K$, the bound scales as $O(\log K)$. Dividing by $K\Delta t$ therefore yields a zero asymptotic rate.
\end{IEEEproof}

Corollary~\ref{cor:terminal} explains the value of trajectory observation. A fixed-dimensional terminal-state observation may suffice for endpoint-level tasks, but it discards the trajectory structure that distinguishes alternative avoidance, passing, and coordination maneuvers. Trajectory observation can instead sustain a nonzero asymptotic mobility information rate when it continues to provide new active modes as the horizon grows.


The spectral mobility gain depends jointly on the maneuver response $\mathbf H(\e^{\mathrm{j}\omega})$ and the output uncertainty spectrum $\mathbf S_Z(\e^{\mathrm{j}\omega})$. A stronger dynamic response to a maneuver therefore does not necessarily improve distinguishability, since the same dynamics may also amplify execution disturbances.



\section{Safety-Aware Maneuver Design}
\label{sec:safety}
We now incorporate workspace boundary, inter-robot separation, and actuator constraints into the mobility information framework.

\subsection{Per-Codeword Safe-Constrained Capacity}
Using the safe maneuver set defined in \eqref{eq:safeset}, we express the MIC under per-codeword safety constraints as
\begin{equation}
C_{M,\max}^{\mathrm{safe}}(T,\eps)=
\sup_{\substack{p(\mathbf{U}):\,\supp(p)\subseteq\cU_{\eps}\\
\mathbb{E}[\mathbf{U}^{\trans}\mathbf{Q}_K\mathbf{U}]\le P_MT}}
I(\mathbf{U};\mathbf{Y}).
\label{eq:exactsafe}
\end{equation}
The formulation in \eqref{eq:exactsafe} requires each maneuver input to satisfy the same individual safety criterion. Direct optimization is difficult because the input distribution must be supported entirely on the safe set. To obtain a tractable formulation, we restrict the maneuver input to a zero-mean Gaussian distribution and impose chance constraints averaged over both maneuver selection and execution uncertainty. These constraints translate physical safety margins into limits on the maneuver-input covariance. However, the resulting Gaussian formulation does not ensure that each individual maneuver is safe.

\subsection{From Physical Chance Constraints to Covariance Constraints}

Given the collision-free nominal trajectories $\bar{\mathbf{p}}_{i,k}$, the position of robot $i$ at time $k$ can be written as
\begin{equation}
\mathbf{p}_{i,k}=\bar{\mathbf{p}}_{i,k}+\mathbf{D}_{i,k}\mathbf{U}+\mathbf{e}_{i,k},
\label{eq:posmap}
\end{equation}
where $\mathbf{D}_{i,k}$ maps the joint maneuver sequence to the displacement from the nominal position. The execution errors $\mathbf{e}_{i,k}$ are jointly zero-mean Gaussian, with marginal covariance $\mathbf{K}_{e,i,k}$. To express the chance constraints in terms of the maneuver resource, we assume throughout this subsection that $\mathbf{U}\sim\mathcal{N}(\mathbf{0},\mathbf{K}_U)$ and that the maneuver input is independent of the execution errors. It follows that $\mathbf{p}_{i,k}$ is Gaussian with mean $\bar{\mathbf{p}}_{i,k}$ and covariance $\mathbf{D}_{i,k}\mathbf{K}_U\mathbf{D}_{i,k}^{\mathsf T}+\mathbf{K}_{e,i,k}$.

We first consider a linear boundary of the admissible workspace, such as a wall. For constraint index $r$, the permitted side is described by
\begin{align}
    \mathbf{a}_r^{\trans}\mathbf{p}_{i,k}\le c_r,
\end{align}
where $\mathbf a_r$ is a unit normal pointing toward the prohibited side and $c_r$ specifies the boundary location. This representation applies to corridor boundaries in two or three dimensions. For an aerial robot, the altitude limit $p_{i,k}^{(z)}\le z_{\max}$ is represented by $\mathbf a_r=[0,0,1]^{\trans}$ and $c_r=z_{\max}$. A separating boundary can also constrain a robot to a prescribed obstacle-free side.

Since the nominal position lies on the permitted side, its distance to the boundary is
\begin{equation}
m_r=c_r-\mathbf{a}_r^{\trans}\bar{\mathbf{p}}_{i,k}>0.
\end{equation}
A violation occurs when the position deviation toward the prohibited side exceeds $m_r$. Substituting \eqref{eq:posmap} expresses this event as $\mathbf{a}_r^{\trans}\mathbf{D}_{i,k}\mathbf{U} +\mathbf{a}_r^{\trans}\mathbf{e}_{i,k}>m_r$. Defining
$\mathbf{S}_r
=\mathbf{D}_{i,k}^{\trans}\mathbf{a}_r
 \mathbf{a}_r^{\trans}\mathbf{D}_{i,k}$ and $\sigma_{r,0}^2=\mathbf{a}_r^{\trans}\mathbf{K}_{e,i,k}\mathbf{a}_r$, we write the total variance of $\mathbf{a}_r^{\trans}\mathbf{D}_{i,k}\mathbf{U} +\mathbf{a}_r^{\trans}\mathbf{e}_{i,k}$ as $v_r = \operatorname{tr}(\mathbf{S}_r\mathbf{K}_U)+\sigma_{r,0}^2$. Here,
$\mathbf{S}_r$ characterizes the contribution of the maneuver-input covariance to position fluctuations along the boundary normal, while $\sigma_{r,0}^2$ is the fixed contribution from execution uncertainty. Together, this variance and the nominal distance $m_r$ determine the Gaussian boundary-crossing probability.

For a risk tolerance $\eps_r$, the following result expresses the probability requirement as an explicit bound on the maneuver-induced variance \cite{ridderhof2022chance}.
\begin{proposition}[Gaussian half-space covariance constraint]
\label{prop:chance}
Under the Gaussian position model above, fix a nominal margin $m_r>0$ and a violation-probability limit $0<\eps_r<1/2$. Define $\kappa_r=\Phi^{-1}(1-\eps_r)$, where $\Phi$ is the standard-normal cumulative distribution function. Then the chance constraint
\begin{equation}
\Pr(\mathbf{a}_r^{\trans}\mathbf{p}_{i,k}>c_r)\le\eps_r
\label{eq:halfchance}
\end{equation}
is equivalent to
\begin{equation}
\Tr(\mathbf{S}_r\mathbf{K}_U)\le b_r(\eps_r),
\label{eq:tracebound}
\end{equation}
where 
\begin{align}
    b_r(\eps_r)=\frac{m_r^2}{\kappa_r^2}-\sigma_{r,0}^2.
\end{align}
Here, the probability is averaged over both maneuver selection and execution uncertainty, and $b_r(\eps_r)$ is the maximum permitted maneuver-induced variance along the boundary normal.
\end{proposition}

\begin{IEEEproof}
A boundary crossing occurs when $\mathbf{a}_r^{\trans} (\mathbf{p}_{i,k}-\bar{\mathbf{p}}_{i,k})>m_r$.
This projected position deviation is zero-mean Gaussian with variance $v_r$. Standardization gives the Gaussian tail condition 
\begin{align}
m_r\ge\kappa_r\sqrt{\operatorname{tr}(\mathbf S_r\mathbf K_U)+\sigma_{r,0}^2}.
\label{eq:scalarcc}
\end{align}
Since both sides are nonnegative, squaring and rearranging yields \eqref{eq:tracebound}.
\end{IEEEproof}

Equation \eqref{eq:scalarcc} bounds the total position variance along the boundary normal. A smaller violation tolerance increases $\kappa_r$ and reduces the variance budget. Execution uncertainty accounts for $\sigma_{r,0}^2$, leaving $b_r(\eps_r)$ for maneuver variation. If $b_r(\eps_r)=0$, no maneuver-induced variance is permitted along that normal. If $b_r(\eps_r)<0$, execution uncertainty alone exceeds the permitted variance, and no $\mathbf K_U$ can satisfy the constraint.

For inter-robot separation, the nominal margin is
\begin{align}
m_r
&=\|\bar{\mathbf{p}}_{i,k}
     -\bar{\mathbf{p}}_{j,k}\|_2-d_{\min}.
\end{align}
Let $\mathbf{n}_{ij,k}
=(\bar{\mathbf{p}}_{i,k}-\bar{\mathbf{p}}_{j,k})/\|\bar{\mathbf{p}}_{i,k}-\bar{\mathbf{p}}_{j,k}\|_2$ denote the nominal separation direction from robot $j$ to robot $i$. To ensure a separation of at least $d_{\min}$, we impose
\begin{equation} 
\|\mathbf{p}_{i,k}-\mathbf{p}_{j,k}\|_2\ge\mathbf{n}_{ij,k}^{\trans}(\mathbf{p}_{i,k}-\mathbf{p}_{j,k})\ge d_{\min}.
\label{eq:pairhalf} 
\end{equation}
Thus, \eqref{eq:pairhalf} is sufficient to ensure the required Euclidean separation \cite{zhu2022decentralized}. With $\mathbf{D}_{ij,k}=\mathbf{D}_{i,k}-\mathbf{D}_{j,k}$, the corresponding covariance constraint is specified by $\mathbf S_r=\mathbf D_{ij,k}^T\mathbf n_{ij,k}\mathbf n_{ij,k}^T \mathbf D_{ij,k}$ and $\sigma_{r,0}^2=\operatorname{Var}
\bigl[\mathbf n_{ij,k}^T(\mathbf e_{i,k}-\mathbf e_{j,k})\bigr]$.
This variance includes the cross-robot execution-error covariance. 

Actuator limits can be treated similarly. Under the linear closed-loop model, each actuator-command component is an affine Gaussian function of $\mathbf U$ and execution uncertainty. Its nominal distance to an upper or lower limit defines the corresponding safety margin. Proposition~\ref{prop:chance} then gives the associated covariance constraint.

Collecting the workspace boundary, pairwise distance, and actuator violation events over the horizon yields $R$ scalar risk events. If their risk budgets satisfy $\sum_{r=1}^{R}\epsilon_r\le\epsilon_T$, Boole's inequality bounds the probability of violating at least one constraint by $\epsilon_T$.

\subsection{Safety-Aware Gaussian Covariance Allocation}
The maneuver covariance $\mathbf K_U$ determines the magnitude and correlation of input variations across robots and time. Physical safety constraints limit the admissible maneuver variation, while observation uncertainty determines how clearly this variation can be distinguished in the observed trajectory. We therefore maximize mobility information subject to the maneuver-resource budget and physical safety margins.

With the maneuver budget and $R$ physical constraints on the maneuver-input covariance, the allocation problem is formulated as
\begin{subequations}
\label{prob:safe}
\begin{align}
\max_{\mathbf{K}_U\succeq\mathbf{0}}\quad
&\frac12\log_2\det\!\left(\mathbf{I}+\mathbf{K}_{Z,K}^{-1/2}\mathbf{G}_K\mathbf{K}_U\mathbf{G}_K^{\trans}\mathbf{K}_{Z,K}^{-1/2}\right) \label{prob:safe-a}\\
\mathrm{s.t.}\quad
&\Tr(\mathbf{Q}_K\mathbf{K}_U)\le P_MT, \label{prob:safe-b}\\
&\Tr(\mathbf{S}_r\mathbf{K}_U)\le b_r(\eps_r),\quad r=1,\ldots,R.\label{prob:safe-c}
\end{align}
\end{subequations}
The bound $b_r(\eps_r)$ accounts for the available physical margin, execution uncertainty, and allocated violation probability. Problem \eqref{prob:safe} is convex: the log-determinant objective is concave in $\mathbf K_U$, the maneuver-resource and safety constraints are affine, and the covariance satisfies $\mathbf K_U\succeq\mathbf 0$. The optimization allocates maneuver variation according to its information gain, resource cost, and use of the available safety margins.

The solution specifies the optimal zero-mean Gaussian maneuver distribution under the average safety constraints. These constraints do not require every Gaussian draw to satisfy the individual risk tolerance; a maneuver with a small selection probability can still have a high execution risk. Replacing the individual safety requirement with an average-risk requirement relaxes the safety constraint, whereas the Gaussian assumption restricts the admissible input distributions. Consequently, the optimum of \eqref{prob:safe} is not generally an upper or lower bound on \eqref{eq:exactsafe}.

\subsection{Water-Filling under Safety Constraints}

To determine when the unconstrained modal interpretation remains applicable, define the cost-normalized maneuver-input covariance and transformed safety matrix as
\begin{align}
\mathbf P&=\mathbf Q_K^{1/2}\mathbf K_U\mathbf Q_K^{1/2},
\\
\widehat{\mathbf S}_r&=
\mathbf Q_K^{-1/2}\mathbf S_r\mathbf Q_K^{-1/2}.
\end{align}
The maneuver budget becomes $\operatorname{tr}(\mathbf P)\le P_MT$, and the safety constraints become
$\operatorname{tr}(\widehat{\mathbf S}_r\mathbf P)\le b_r$. Each $\widehat{\mathbf S}_r$ is positive semidefinite and rank one because it originates from a scalar half-space event. The mobility modes are still the eigenvectors of $\widetilde{\mathbf{G}}_K^{\trans}\widetilde{\mathbf{G}}_K$. We next examine whether the transformed safety matrices $\widehat{\mathbf{S}}_r$ can be diagonalized in the same basis, allowing the matrix optimization to reduce to resource allocation across these modes.

\begin{proposition}[Mobility-mode allocation under safety constraints]
Let $\mathbf V$ diagonalize $\widetilde{\mathbf G}_K^{\trans}\widetilde{\mathbf G}_K$, with eigenvalues $\lambda_\ell$. If every $\widehat{\mathbf S}_r$ is diagonal in this basis, an optimum of \eqref{prob:safe} can be chosen in the same mobility-mode basis:
\begin{equation}
\label{eq:pstar}
\mathbf{P}^{\star}
=\mathbf{V}\diag(p_1^{\star},p_2^{\star},\ldots)
\mathbf{V}^{\trans}.
\end{equation}
For each positive-gain mode, define $d_{r\ell}=\mathbf v_\ell^T\widehat{\mathbf S}_r\mathbf v_\ell$.
The safety constraint becomes $\sum_\ell d_{r\ell}p_\ell\le b_r$, and the optimal allocation is
\begin{equation}
p_\ell^\star=
\left[
\frac{1}{
2\ln2\left(\mu_0+\sum_{r=1}^{R}\mu_r d_{r\ell}\right)}
-\frac{1}{\lambda_\ell}
\right]^+,
\label{eq:multiprice}
\end{equation}
where the nonnegative multipliers enforce the maneuver budget and the active safety constraints.

If the safety constraints further reduce to independent modal limits $0\le p_\ell\le p_\ell^{\mathrm{safe}}$, then
\begin{equation}
p_\ell^\star=
\min\left\{
p_\ell^{\mathrm{safe}},
\left(\nu-\frac{1}{\lambda_\ell}\right)^+
\right\},
\label{eq:capped}
\end{equation}
where $\nu$ is chosen to meet the available total maneuver budget.
\end{proposition}

\begin{IEEEproof}
Let $\mathbf X=\mathbf V^{\trans}\mathbf P\mathbf V\succeq\mathbf 0$. By invariance of the trace, the maneuver-resource constraint is $\operatorname{tr}(\mathbf X)\le P_MT$. Moreover,
\begin{align}
\operatorname{tr}(\widehat{\mathbf S}_r\mathbf P)
=
\operatorname{tr}\!\left(
\mathbf V^{\trans}\widehat{\mathbf S}_r\mathbf V\mathbf X
\right)
=
\sum_{\ell=1}^{L_K}d_{r\ell}x_{\ell\ell},
\end{align}
so every safety constraint depends only on the diagonal entries of $\mathbf X$. Using Sylvester's determinant identity and Hadamard's inequality, an optimum can be chosen with $\mathbf X=\diag(p_1,\ldots,p_L)$, which gives \eqref{eq:pstar}.

With this diagonal representation, the matrix optimization reduces to allocating a nonnegative maneuver resource $p_\ell$ to each mobility mode. The reduced problem maximizes \eqref{eq:parallel} subject to $\sum_{\ell=1}^{L_K}p_\ell\le P_MT$ and $\sum_{\ell=1}^{L_K}d_{r\ell}p_\ell\le b_r$. The Karush--Kuhn--Tucker (KKT) conditions then yield the multi-price allocation in \eqref{eq:multiprice}. In the scalar half-space model of Section~V-B, every $\widehat{\mathbf S}_r$ is rank one. Therefore, such a matrix has at most one positive diagonal entry and imposes an individual cap on one mode. Hence, the simpler capped water-filling rule \eqref{eq:capped} applies under the independent maneuver-cost limits $0\le p_\ell\le p_\ell^{\mathrm{safe}}$.
\end{IEEEproof}

The allocation rule balances the observable gain of each mobility mode against its resource cost and its use of active safety margins. The gain $\lambda_\ell$ describes how clearly a unit of normalized maneuver variation can be observed, while $\sum_r\mu_r d_{r\ell}$ accounts for its effect on the active physical constraints. A highly observable mode can therefore receive little or no allocation when it strongly excites a restrictive constraint. A mode with a smaller observable gain can remain useful if it consumes less of the available safety margin.

The alignment between the physical constraints and the mobility modes determines the allocation structure. When the transformed safety matrices share the mobility-mode basis, the scalar allocation in Proposition 3 is optimal. For general coupled constraints, the full covariance optimization also determines the correlations between mobility modes.

\subsection{Finite Per-Codeword Safe Mobility Codebook}
\label{sec:codebook}
A finite joint maneuver codebook consists of a set $\mathcal C=\{\mathbf u^{(1)},\ldots,\mathbf u^{(M)}\}$ of deterministic joint maneuver sequences, with each $\mathbf u^{(m)}$ serving as a codeword. Per-codeword safety requires each sequence to satisfy the prescribed risk threshold over execution uncertainty. The solution \(\mathbf K_U^\star\) of the covariance allocation problem \eqref{prob:safe} specifies a continuous Gaussian maneuver distribution rather than a finite executable set. We use \(\mathcal N(\mathbf 0,\mathbf K_U^\star)\) as a proposal distribution to generate candidate joint fleet maneuvers, screen each candidate for safety, and select codewords based on their trajectory separation.

For a fixed candidate $\mathbf{u}$, uncertainty arises only from execution error. For each scalar safety constraint indexed by $r$, the corresponding safety margin is written as
\begin{equation}
m_r+\mathbf g_r^{\trans}\mathbf u+\xi_r,
\end{equation}
where $\mathbf g_r\triangleq-\mathbf D_{i,k}^{\trans}\mathbf a_r$ characterizes the maneuver-induced change in the margin and $\xi_r\sim\mathcal N(0,\sigma_{r,0}^2)$ represents the projected execution uncertainty. Both the workspace boundary constraints and the pairwise-distance constraint in \eqref{eq:pairhalf} admit this safety-margin representation. A violation occurs when the margin is negative, yielding
\begin{equation}
\begin{aligned}
\Pr(\mathcal E_r\mid\mathbf u)
=
\Phi\!\left(
-\frac{m_r+\mathbf g_r^{\trans}\mathbf u}{\sigma_{r,0}}
\right).
\end{aligned}
\label{eq:conditionalcert}
\end{equation}
Each maneuver candidate is screened against all safety events and retained only if it satisfies
\begin{align}
\rho_{\mathrm{cb}}(\mathbf u)
&=
\sum_{r=1}^{R}\Pr(\mathcal E_r\mid\mathbf u)
\le \eps_{\mathrm{cb}},
\label{eq:rhocb}
\end{align}
where $\epsilon_{\mathrm{cb}}$ is the per-codeword risk threshold. By Boole's inequality, this condition guarantees
\begin{equation}
\Pr\!\left(\bigcup_{r=1}^{R}\mathcal E_r\mid\mathbf u\right)
\le
\epsilon_{\mathrm{cb}}.
\label{eq:cb_union}
\end{equation}
In addition to the probabilistic safety test, each candidate must satisfy the deterministic maneuver-cost constraint.

For two maneuver candidates indexed by $m$ and $m'$, define their noise-whitened trajectory distance as
\begin{equation}
d_{m,m'}=\left\|\mathbf{K}_{Z,K}^{-1/2}\mathbf{G}_K\big(\mathbf{u}^{(m)}-\mathbf{u}^{(m')}\big)\right\|_2.
\label{eq:mahaldist}
\end{equation}
This quantity is the Mahalanobis distance between the mean trajectory observations induced by the two candidates. A larger $d_{m,m'}$ makes the corresponding maneuvers easier for the mobility monitor to distinguish. Starting from an initial codeword, we construct the codebook sequentially by selecting the remaining candidate with the largest distance to its nearest retained codeword, as described in Algorithm~\ref{alg:codebook}.

For the resulting codebook, let $J\sim\mathrm{Unif}\{1,\ldots,M\}$ and let the realization $J=m$ select $\mathbf u^{(m)}$. Then
\begin{equation}
\mathbf Y\mid J=m\sim\mathcal N\!\left(\mathbf G_K\mathbf u^{(m)},\mathbf K_{Z,K}\right).
\end{equation}
Hence, under equal codeword priors, the mobility monitor applies maximum-likelihood decoding. For a selected index $m$ and a competing index $m'$, define
\begin{equation}
\bm{\delta}_{m,m'}\triangleq\mathbf{G}_K\big(\mathbf{u}^{(m)}-\mathbf{u}^{(m')}\big).
\end{equation}
Let $\varphi_{\mathbf K_{Z,K}}$ denote the zero-mean Gaussian density with covariance $\mathbf K_{Z,K}$. The mutual information achieved by the finite codebook is then
\begin{align}
I(J;\mathbf{Y})=\frac{1}{M}\sum_{m=1}^{M}\mathbb{E}_{\mathbf{Z}_K}
\left[
\log_2\frac{M\varphi_{\mathbf{K}_{Z,K}}(\mathbf{Z}_K)}
{\sum_{m'=1}^{M}\varphi_{\mathbf{K}_{Z,K}}(\mathbf{Z}_K+\bm{\delta}_{m,m'})}
\right].
\label{eq:mixtureMI}
\end{align}
This expression evaluates the actual finite codebook, whose input distribution is no longer Gaussian following safety screening and codeword selection.

\begin{algorithmblock}{Safe Joint Mobility Codebook Construction}
\label{alg:codebook}

\noindent\textbf{Input:} $\mathbf K_U^\star$, $\mathbf G_K$, $\mathbf K_{Z,K}$,$\mathbf Q_K$, $P_M T$, $\{m_r,\mathbf g_r,\sigma_{r,0}\}_{r=1}^{R}$, per-codeword risk threshold $\eps_{\mathrm{cb}}$, candidate count $N_c$, and target codebook size $M_{\mathrm{tar}}$.

\begin{enumerate}
\item Generate $N_c$ joint maneuver candidates from $\mathcal N(\mathbf 0,\mathbf K_U^\star)$.
\item Reject every candidate $\mathbf u$ with $\mathbf u^{\trans}\mathbf Q_K\mathbf u>P_M T$.
\item For each remaining candidate, compute $\rho_{\mathrm{cb}}(\mathbf u)$ and retain only those satisfying \eqref{eq:rhocb}. If no candidates remain, return $\mathcal C=\emptyset$ and stop.
\item Among the remaining candidates, initialize the codebook $\mathcal C$ with $\mathbf u^\star
=
\arg\max_{\mathbf u}
\left\|
\mathbf K_{Z,K}^{-1/2}\mathbf G_K\mathbf u
\right\|_2,
\qquad
\mathcal C=\{\mathbf u^\star\}.
$ 

\item While $|\mathcal C|<M_{\mathrm{tar}}$ and unselected candidates remain, add the candidate $\mathbf u^\star$ satisfying
\[
\mathbf u^\star
=
\arg\max_{\mathbf u}
\min_{\mathbf u^{(m')}\in\mathcal C}
\left\|
\mathbf K_{Z,K}^{-1/2}\mathbf G_K
\bigl(\mathbf u-\mathbf u^{(m')}\bigr)
\right\|_2,
\]
and update
$\mathcal C\leftarrow\mathcal C\cup\{\mathbf u^\star\}$.

\end{enumerate}

\noindent\textbf{Output:}
The finite joint maneuver codebook $\mathcal C$ with actual size $M=|\mathcal C|\le M_{\mathrm{tar}}$.

\end{algorithmblock}

\begin{proposition}[Finite-codebook safety and identification bound]
Suppose that the scalar Gaussian safety-margin model above holds for all $r=1,\ldots,R$. Every codeword $\mathbf u^{(m)}$, $m=1,\ldots,M$, generated by Algorithm~\ref{alg:codebook} satisfies the maneuver-resource constraint and
\begin{equation}
\Pr\!\left(\bigcup_{r=1}^{R}\mathcal E_r\,\middle|\,\mathbf u^{(m)}\right)\le\eps_{\mathrm{cb}}.
\end{equation}
For $M\ge 2$, define the minimum pairwise trajectory distance
\begin{equation}
d_{\min}^{\mathcal C}
=
\min_{m\neq m'} d_{m,m'}.
\end{equation}
Under equal codeword priors, the maximum-likelihood decoder satisfies
\begin{equation}
P_e
\le
(M-1)
Q\!\left(
\frac{d_{\min}^{\mathcal C}}{2}
\right).
\label{eq:codebook_error_bound}
\end{equation}
\end{proposition}

\begin{IEEEproof}
The safety result follows directly from \eqref{eq:cb_union} and the screening step of Algorithm~\ref{alg:codebook}, while Step 2 ensures that the maneuver-resource constraint is satisfied. After the observation noise is whitened, conditional on the selection of maneuver sequence \(\mathbf u^{(m)}\), the probability that a competing codeword $m'$ has a higher likelihood than $m$ is $Q(d_{m,m'}/2)$. Applying a union bound over the $M-1$ competing codewords gives \eqref{eq:codebook_error_bound}.
\end{IEEEproof}

For a codebook with $M\ge2$, define
\begin{equation}
\bar\eta_{\mathcal C}
=\min\left\{(M-1)Q\!\left(
\frac{d_{\min}^{\mathcal C}}{2}\right),
1-\frac{1}{M}\right\}.
\label{eq:mic_library_error_bound}
\end{equation}
The first term follows from Proposition~4, and the second follows because maximum-likelihood decoding with equal priors performs at least as well as random guessing. Hence, $P_e\le\bar\eta_{\mathcal C}\le 1-1/M$. Applying Fano's inequality and using the feasibility of the constructed codebook yields
\begin{align}
&\log_2 M
-h_2(\bar\eta_{\mathcal C})
-\bar\eta_{\mathcal C}\log_2(M-1)
\nonumber\\
&\qquad\le I(J;\mathbf Y)
\le C_{M,\max}^{\rm safe}(T,\eps_{\mathrm{cb}})
\le C_M^{\rm G}(T).
\label{eq:mic_library_capacity_link}
\end{align}
The first upper bound holds because every retained codeword satisfies the per-codeword risk requirement and the maneuver-resource constraint. The uniform distribution over the codebook is therefore a feasible input distribution for $C_{M,\max}^{\rm safe}(T,\eps_{\mathrm{cb}})$. Removing the safety requirement enlarges the input class, yielding $C_{M,\max}^{\rm safe}(T,\eps_{\mathrm{cb}}) \le C_M^{\rm G}(T)$. Thus, \eqref{eq:mic_library_capacity_link} relates the information achieved by the finite codebook to an error-based achievable lower bound and the two capacity benchmarks. Since safety screening and distance-based selection modify the Gaussian proposal distribution, we evaluate the actual finite-codebook mutual information using \eqref{eq:mixtureMI}.

\section{Aerial-Robot Case Study}
\label{sec:simulation}
\begin{figure}[t]
    \centering
    \includegraphics[width=0.50\textwidth]{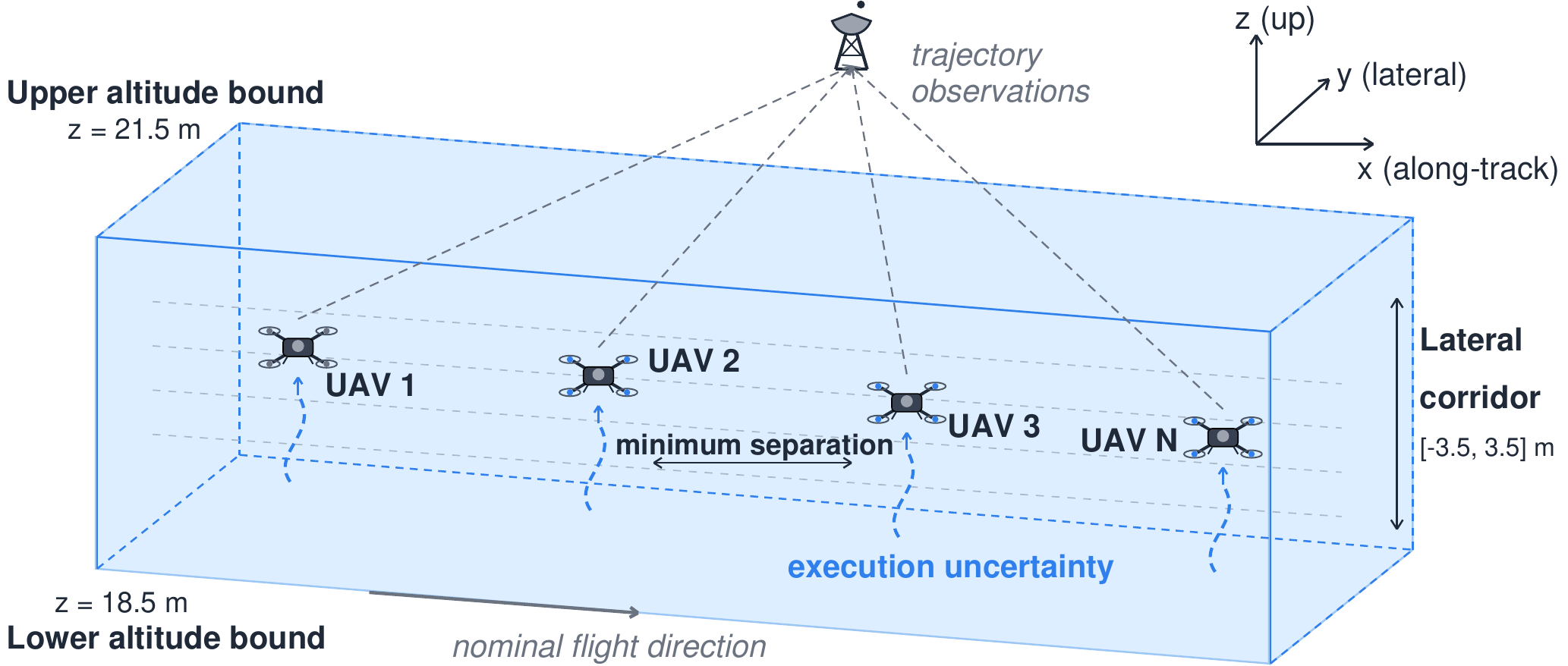}
    \caption{Aerial-robot evaluation scenario. The mobility monitor observes position trajectories in a corridor with altitude and inter-robot separation constraints.}
    \label{fig:simulation_scenario}
\end{figure}

We evaluate the proposed framework using the aerial-robot fleet shown in Fig.~\ref{fig:simulation_scenario}. For robot $i$, the state consists of position and velocity, i.e., $\mathbf x_{i,k}
=\left[\mathbf p_{i,k}^{\trans}, 
\boldsymbol{\upsilon}_{i,k}^{\trans}\right]^{\trans}$, with acceleration as the control input. The control update interval is $\Delta t=0.25$ s, and acceleration is held constant within each interval. A fixed stabilizing feedback controller governs the nominal closed-loop motion, while the intentional maneuver input is an additional exogenous acceleration sequence. Spatially correlated acceleration disturbances represent environmental uncertainty shared by the robots. The resulting model is
\begin{align}
\mathbf x_{i,k+1} &= \mathbf A\mathbf x_{i,k} + \mathbf B\mathbf u_{i,k} + \mathbf w_{i,k},
\label{eq:sim_dynamics}\\
\mathbf y_{i,k} &= \mathbf C\mathbf x_{i,k+1} + \mathbf v_{i,k},
\label{eq:sim_observation}
\end{align}
where
\begin{equation}
\mathbf B
=
\begin{bmatrix}
\frac{\Delta t^2}{2}\mathbf I_3\\
\Delta t\,\mathbf I_3
\end{bmatrix},
\qquad
\mathbf C
=
\begin{bmatrix}
\mathbf I_3 & \mathbf 0
\end{bmatrix}.
\end{equation}

The aerial robots fly in parallel at $2$ m/s through a bounded corridor. Their lateral lanes span $[-2.75,2.75]$ m, and their altitudes alternate between $19.85$ and $20.15$ m. Nominal relative-position vectors define the pairwise half-spaces in \eqref{eq:pairhalf}. The maneuver design includes corridor, altitude, and pairwise-distance constraints, with the latter constructed using the relative-position directions in \eqref{eq:pairhalf}.

Table~\ref{tab:parameters} summarizes the simulation parameters. Unless otherwise specified, the same configuration is used in all experiments.

\begin{table}[t]
\caption{Simulation Setup}
\label{tab:parameters}
\centering
\footnotesize
\begin{tabular}{p{2.35cm}p{0.9cm}p{4cm}}
\toprule
Parameter & Symbol & Value / source \\
\midrule
Fleet size & $N$ & $1$--$6$ robots \\
Control update interval & $\Delta t$ & $0.25$ s \\ 
Corridor / altitude & -- & $|y|\le3.5$ m; $18.5\le z\le21.5$ m \\
Minimum inter-robot distance & $d_{\min}$ & $2.5$ m (risk) \\
Acceleration-disturbance standard deviation$^{a}$ & -- & $(0.08,0.08,0.05)$ m/s$^2$ \\
Position-observation standard deviation & -- & $(0.15,0.15,0.10)$ m \\
Cross-robot correlation$^{a}$ & $\rho_w$ & $0.15$ \\
Maneuver budget per axis and instant & -- & $0.04$ (m/s$^2$)$^2$ \\
\bottomrule
\multicolumn{3}{p{8.0cm}}{\scriptsize
$^{a}$The acceleration disturbances represent wind-induced motion
perturbations in the aerial-robot case study. Their standard deviations
are specified before propagation through the acceleration-input blocks.
$\rho_w$ describes the correlation between different robots' disturbances.}
\end{tabular}
\end{table}

The simulations evaluate three aspects of the proposed MIC framework. First, we examine convergence from the finite-horizon gain distribution to its stationary spectral characterization and compare full-trajectory and terminal-only observations. Second, we assess how physical safety constraints affect mobility information by comparing equal allocation, unconstrained water-filling in \eqref{eq:wf}, the general safety-aware solution of \eqref{prob:safe}, and a modal-cap allocation based on capped water-filling. We also evaluate the finite maneuver codebook constructed by Algorithm~\ref{alg:codebook} in terms of achieved mutual information, maneuver-identification error, and per-codeword safety. Third, we examine fleet-wide and per-robot information as more aerial robots share the same corridor.

\begin{figure}[!t]
  \centering
  \subfigure[]{%
    \includegraphics[width=0.45\textwidth]{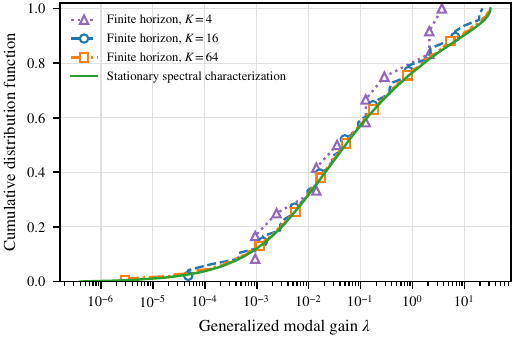}%
    \label{fig:spectral:gains}%
  }\hfill
  \subfigure[]{%
    \includegraphics[width=0.45\textwidth]{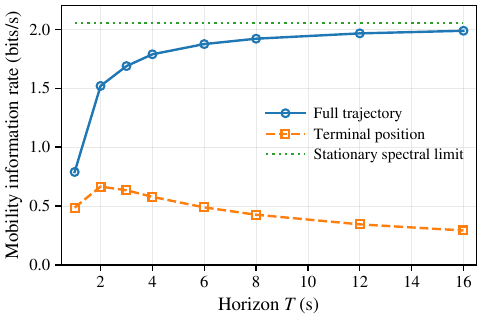}%
    \label{fig:spectral:rates}%
  }
  \caption{Spectral convergence. (a) Finite-horizon and stationary generalized-gain distributions. (b) Information rates with full-trajectory and terminal-state observations.}
  \label{fig:spectral}
\end{figure}

We first examine the long-horizon behavior predicted by Theorem~\ref{thm:spectral} and the mobility information retained through full-trajectory observation. In Fig.~\ref{fig:spectral:gains}, the finite-horizon generalized-gain distributions approach the stationary spectral characterization as $K$ increases from $4$ to $64$. At $K=4$, the finite-horizon distribution differs noticeably from its stationary counterpart. This difference decreases as the horizon grows, indicating that the spectral model captures the asymptotic mobility modes. Fig.~\ref{fig:spectral:rates} compares the information rates obtained with full-trajectory and terminal-only observations. At $K=64$ ($T=16$ s), the full-trajectory rate is close to the stationary spectral limit, whereas terminal-only observation incurs a substantial rate loss, consistent with Corollary~\ref{cor:terminal}. These trends are consistent with the sustained rate characterized by Theorem 2 and the terminal-observation scaling in Corollary 2, demonstrating the value of retaining trajectory history for mobility monitoring.


\begin{figure}[!t]
  \centering
  \subfigure[]{%
    \includegraphics[width=0.45\textwidth]{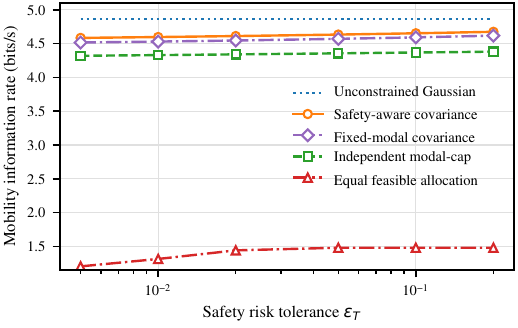}%
    \label{fig:safetytradeoff}%
  }\hfill
  \subfigure[]{%
    \includegraphics[width=0.45\textwidth]{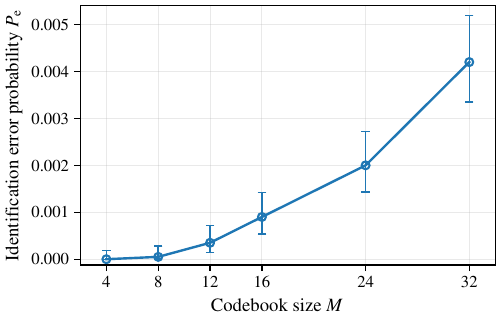}%
    \label{fig:codebook:error}%
  }
  \caption{Safety-aware maneuver allocation and finite-codebook performance. (a) Safety--information tradeoff under constraints. (b) ML identification error versus codebook size $M$ (error bars indicate two-sided 95\% Clopper--Pearson confidence intervals). }
  \label{fig:safety_codebook}
\end{figure}

Fig.~\ref{fig:safetytradeoff} compares how maneuver allocation and the representation of physical constraints affect mobility information.  The unconstrained Gaussian solution provides a reference without physical safety restrictions. The safety-aware design optimizes the full maneuver covariance, whereas the fixed-modal design restricts the cost-normalized covariance to $\mathbf P=\mathbf V\operatorname{diag}(p_\ell)\mathbf V^{\mathsf T}$ while retaining the original coupled safety inequalities. The independent modal-cap design applies capped water-filling under separate modal limits. Equal feasible allocation uses $\mathbf K_U=\alpha\mathbf I$, with the largest $\alpha$ satisfying the maneuver-resource and safety constraints. The safety-aware covariance design achieves a substantially higher information rate than equal feasible allocation and retains most of the unconstrained reference. The fixed-modal design remains close to the full covariance solution, showing that allocation in the mobility-mode basis captures most of the available gain. Independent modal caps yield a lower rate because they restrict the tradeoffs permitted by the coupled safety inequalities. These results show that observable mobility gains and the structure of physical safety constraints jointly determine effective maneuver allocation.

Fig.~\ref{fig:codebook:error} evaluates the identification reliability of the finite joint maneuver codebooks constructed by Algorithm 1.  Across the tested codebook sizes, the empirical ML identification error remains below $0.5\%$. For the $32$-codeword design, evaluating the finite-codebook mutual information \eqref{eq:mixtureMI} gives approximately $4.98$ bits, close to the $5$-bit entropy of a uniformly selected codeword. The trajectory observations therefore retain nearly all the information carried by the selected maneuver index. Increasing the codebook size introduces more maneuver alternatives within the same resource-constrained observation space and increases the identification error, illustrating the tradeoff between maneuver diversity and identification reliability.

\begin{figure}[!t]
  \centering
  \subfigure[]{%
    \includegraphics[width=0.45\textwidth]{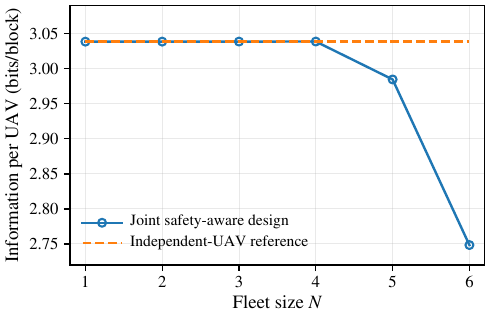}%
    \label{fig:fleet:peruav}%
  }\hfill
  \subfigure[]{%
    \includegraphics[width=0.45\textwidth]{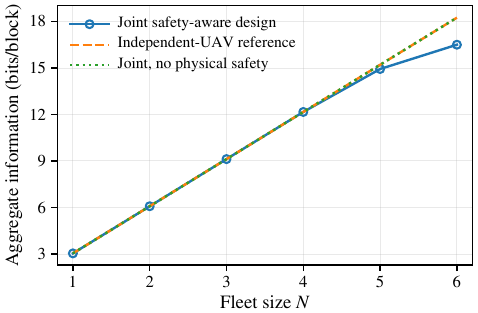}%
    \label{fig:fleet:aggregate}%
  }
  \caption{Aerial-robot fleet scaling in a fixed corridor. (a) Mobility information per robot. (b) Aggregate mobility information.}
  \label{fig:fleet}
\end{figure}

Finally, Figure 6 identifies when additional robots begin to consume one another’s maneuver freedom in the fixed corridor. With the per-robot, per-step maneuver budget held constant, information per robot remains approximately constant up to \(N=4\) and decreases when the pairwise-distance constraints become active from \(N=5\). The design without physical safety constraints remains close to the independent-robot reference, indicating that the reduction is primarily associated with the increasingly restrictive spatial safety margins. Aggregate mobility information continues to grow as additional robots introduce new joint maneuver dimensions
At $N=6$, the safety-aware design retains about $90\%$ of the independent-robot information, and its aggregate information exceeds five times the single-robot information. These results distinguish the fleet-wide information gained by adding robots from the maneuver freedom available to each robot. MIC captures both effects and identifies how shared-space constraints change the information supported by a given fleet size.

\section{Conclusion}
\label{sec:discussion}
This paper established a mobility information capacity framework for quantifying observable maneuver freedom in multi-robot systems. Cost- and uncertainty-normalized mobility modes connect the fleet dynamics to the distinguishability of its trajectory observations. The stationary spectral characterization identifies when maneuver information grows proportionally with the observation horizon, while the terminal-observation analysis establishes conditions under which a fixed-dimensional endpoint cannot sustain a nonzero information rate. Physical safety margins enter maneuver design through coupled covariance constraints, determining the allocation across mobility modes and the role of correlations between them.

Individual risk screening and noise-whitened trajectory separation translate the covariance design into finite joint maneuver codebooks with explicit execution-risk and identification-error bounds. The aerial-robot study demonstrates the results for full-trajectory observation, safety-aware resource allocation, and fleet scaling in a shared corridor. Together, these results provide a quantitative basis for selecting maneuver variation according to its observability, resource cost, and use of physical safety margins.

%
\bibliographystyle{IEEEtran}
\bibliography{references}

@article{shannon1948,
  author  = {C. E. Shannon},
  title   = {A Mathematical Theory of Communication},
  journal = {Bell System Technical Journal},
  volume  = {27},
  pages   = {379--423, 623--656},
  year    = {1948}
}

@article{shah2026multiagent,
  title         = {Multi-Agent Empowerment and Emergence of Complex Behavior in Groups},
  author        = {Shah, Tristan and Nemenman, Ilya and Polani, Daniel and Tiomkin, Stas},
  year          = {2026},
  eprint        = {2604.21155},
  archivePrefix = {arXiv},
  primaryClass  = {cs.AI},
  url           = {https://arxiv.org/abs/2604.21155}
}

@article{yuan2026mobility,
  author  = {Weijie Yuan and Fan Liu and Shuangyang Li and Lin Zhou and Pingzhi Fan},
  title   = {Mobility Information Capacity in the Sky: A {Gaussian} Channel Perspective},
  journal = {arXiv preprint arXiv:2609.10436},
  year    = {2026},
  url     = {https://arxiv.org/abs/2609.10436}
}

@book{cover2006,
  author    = {T. M. Cover and J. A. Thomas},
  title     = {Elements of Information Theory},
  edition   = {2nd},
  publisher = {Wiley},
  address   = {Hoboken, NJ, USA},
  year      = {2006}
}

@article{gutierrez2008,
  author  = {Jes{\'u}s Guti{\'e}rrez-Guti{\'e}rrez and Pedro M. Crespo},
  title   = {Asymptotically Equivalent Sequences of Matrices and
             Hermitian Block Toeplitz Matrices With Continuous Symbols:
             Applications to {MIMO} Systems},
  journal = {IEEE Transactions on Information Theory},
  volume  = {54},
  number  = {12},
  pages   = {5671--5680},
  month   = dec,
  year    = {2008},
  doi     = {10.1109/TIT.2008.2006401}
}

@article{hoffman1953variation,
  title={The variation of the spectrum of a normal matrix},
  author={Hoffman, Alan J and Wielandt, Helmut W},
  journal={Duke Mathematical Journal},
  volume={20},
  number={1},
  pages={37--39},
  year={1953},
  publisher={Duke University Press}
}

@article{delsarte1991tridiagonal,
  title={Tridiagonal approach to the algebraic environment of Toeplitz matrices, part I: Basic results},
  author={Delsarte, Philippe and Genin, Yves},
  journal={SIAM Journal on Matrix Analysis and Applications},
  volume={12},
  number={2},
  pages={220--238},
  year={1991},
  publisher={SIAM}
}

@article{gazzah2001asymptotic,
  title={Asymptotic eigenvalue distribution of block Toeplitz matrices and application to blind SIMO channel identification},
  author={Gazzah, Houcem and Regalia, Phillip A and Delmas, J-P},
  journal={IEEE transactions on information theory},
  volume={47},
  number={3},
  pages={1243--1251},
  year={2001},
  publisher={IEEE}
}

@inproceedings{klyubin2005,
  author    = {A. S. Klyubin and D. Polani and C. L. Nehaniv},
  title     = {Empowerment: A Universal Agent-Centric Measure of Control},
  booktitle = {Proc. IEEE Congress on Evolutionary Computation},
  volume    = {1},
  pages     = {128--135},
  year      = {2005}
}

@article{tiomkin2024,
  author  = {S. Tiomkin and I. Nemenman and D. Polani and N. Tishby},
  title   = {Intrinsic Motivation in Dynamical Control Systems},
  journal = {PRX Life},
  volume  = {2},
  pages   = {033009},
  year    = {2024},
  doi     = {10.1103/PRXLife.2.033009}
}

@article{nair2004,
  author  = {G. N. Nair and R. J. Evans and I. M. Y. Mareels and W. Moran},
  title   = {Topological Feedback Entropy and Nonlinear Stabilization},
  journal = {IEEE Transactions on Automatic Control},
  volume  = {49},
  number  = {9},
  pages   = {1585--1597},
  month   = sep,
  year    = {2004}
}

@article{jin2025tmc,
  author  = {H. Jin and J. Wu and W. Yuan and F. Liu and Y. Cui},
  title   = {Co-Design of Sensing, Communications, and Control for Low-Altitude Wireless Networks},
  journal = {IEEE Transactions on Mobile Computing},
  volume  = {24},
  number  = {11},
  pages   = {12035--12048},
  month   = nov,
  year    = {2025},
  doi     = {10.1109/TMC.2025.3581616}
}

@article{wang2021airtraffic,
  author  = {Z. Wang and D. Delahaye and J.-L. Farges and S. Alam},
  title   = {Air Traffic Assignment for Intensive Urban Air Mobility Operations},
  journal = {Journal of Aerospace Information Systems},
  volume  = {18},
  number  = {11},
  pages   = {860--875},
  month   = nov,
  year    = {2021},
  doi     = {10.2514/1.I010954}
}

@article{sun2024multiuavmec,
  author  = {G. Sun and Y. Wang and Z. Sun and Q. Wu and J. Kang and D. Niyato and V. C. M. Leung},
  title   = {Multi-Objective Optimization for Multi-{UAV}-Assisted Mobile Edge Computing},
  journal = {IEEE Transactions on Mobile Computing},
  volume  = {23},
  number  = {12},
  pages   = {14803--14820},
  month   = dec,
  year    = {2024},
  doi     = {10.1109/TMC.2024.3446819}
}

@article{wei2026airfogsim,
  author  = {Z. Wei and B. Li and R. Zhang and X. Cheng and L. Yang},
  title   = {{AirFogSim}: A Light-Weight and Modular Simulator for {UAV}-Integrated Vehicular Fog Computing},
  journal = {IEEE Transactions on Mobile Computing},
  volume  = {25},
  number  = {5},
  pages   = {6755--6766},
  month   = may,
  year    = {2026},
  doi     = {10.1109/TMC.2025.3641373}
}

@article{mozaffari2017,
  author  = {M. Mozaffari and W. Saad and M. Bennis and M. Debbah},
  title   = {Mobile Unmanned Aerial Vehicles ({UAVs}) for Energy-Efficient Internet of Things Communications},
  journal = {IEEE Transactions on Wireless Communications},
  volume  = {16},
  number  = {11},
  pages   = {7574--7589},
  month   = nov,
  year    = {2017},
  doi     = {10.1109/TWC.2017.2751045}
}

@article{meng2023,
  author  = {K. Meng and X. He and Q. Wu and D. Li},
  title   = {Multi-{UAV} Collaborative Sensing and Communication: Joint Task Allocation and Power Optimization},
  journal = {IEEE Transactions on Wireless Communications},
  volume  = {22},
  number  = {6},
  pages   = {4232--4246},
  month   = jun,
  year    = {2023},
  doi     = {10.1109/TWC.2022.3224143}
}

@article{wu2024isac,
  author  = {J. Wu and W. Yuan and L. Hanzo},
  title   = {When {UAVs} Meet {ISAC}: Real-Time Trajectory Design for Secure Communications},
  journal = {IEEE Transactions on Vehicular Technology},
  volume  = {72},
  number  = {12},
  pages   = {16766--16771},
  month   = dec,
  year    = {2023},
  doi     = {10.1109/TVT.2023.3290033}
}

@article{motlagh2016,
  author  = {N. H. Motlagh and T. Taleb and O. Arouk},
  title   = {Low-Altitude Unmanned Aerial Vehicles-Based Internet of Things Services: Comprehensive Survey and Future Perspectives},
  journal = {IEEE Internet of Things Journal},
  volume  = {3},
  number  = {6},
  pages   = {899--922},
  month   = dec,
  year    = {2016},
  doi     = {10.1109/JIOT.2016.2612119}
}

@article{garg2024safe,
  author  = {Garg, Kunal and Zhang, Songyuan and So, Oswin and Dawson, Charles and Fan, Chuchu},
  title   = {Learning Safe Control for Multi-Robot Systems: Methods, Verification, and Open Challenges},
  journal = {Annual Reviews in Control},
  volume  = {57},
  pages   = {100948},
  year    = {2024},
  doi     = {10.1016/j.arcontrol.2024.100948}
}

@article{levy2025representation,
  author  = {Levy, Andrew and Allievi, Alessandro G. and Konidaris, George},
  title   = {Representation Learning and Skill Discovery with Empowerment},
  journal = {Reinforcement Learning Journal},
  volume  = {6},
  pages   = {2762--2787},
  year    = {2025},
  url     = {https://rlj.org/papers/levy2025representation}
}

@article{bolten2022toeplitz,
  author  = {Bolten, Matthias and Ekstr{\"o}m, Sven-Erik and Furci, Isabella and Serra-Capizzano, Stefano},
  title   = {{Toeplitz} Momentary Symbols: Definition, Results, and Limitations in the Spectral Analysis of Structured Matrices},
  journal = {Linear Algebra and its Applications},
  volume  = {651},
  pages   = {51--82},
  year    = {2022},
  doi     = {10.1016/j.laa.2022.06.017}
}

@article{ridderhof2022chance,
  author  = {Ridderhof, Jack and Tsiotras, Panagiotis},
  title   = {Chance-Constrained Covariance Steering in a {Gaussian} Random Field via Successive Convex Programming},
  journal = {Journal of Guidance, Control, and Dynamics},
  volume  = {45},
  number  = {4},
  pages   = {599--610},
  year    = {2022},
  doi     = {10.2514/1.G005941}
}

@article{zhu2022decentralized,
  author  = {Zhu, Hai and Brito, Bruno and Alonso-Mora, Javier},
  title   = {Decentralized Probabilistic Multi-Robot Collision Avoidance Using Buffered Uncertainty-Aware {Voronoi} Cells},
  journal = {Autonomous Robots},
  volume  = {46},
  number  = {2},
  pages   = {401--420},
  year    = {2022},
  doi     = {10.1007/s10514-021-10029-2}
}

\end{document}


\title{Supplemental Material for\\
Mobility Information Capacity of Multi-Robot Systems: Safety-Aware Maneuver Design}
\author{Weijie Yuan,~\IEEEmembership{Senior Member,~IEEE}}
\maketitle

\noindent This supplemental material contains the proof of the long-horizon
capacity theorem in the main manuscript. Section, figure, theorem, and
unprefixed equation references refer to the main manuscript. Equations in
this supplement are numbered with the prefix S.

\setcounter{equation}{0}
\renewcommand{\theequation}{S\arabic{equation}}

\appendices
\section{Proof of the Long-Horizon Capacity Theorem}
\label{app:spectral-proof}

\subsection{From Finite-Horizon Gains to Spectral Gains}

For a horizon of $K$ time steps, define the normalized Gram matrix
\begin{equation}
\mathbf M_K
\triangleq
\widetilde{\mathbf G}_K^{\trans}
\widetilde{\mathbf G}_K
=
\mathbf Q_K^{-1/2}
\mathbf G_K^{\trans}
\mathbf K_{Z,K}^{-1}
\mathbf G_K
\mathbf Q_K^{-1/2}.
\label{eq:app_MK}
\end{equation}
Denote all $m_uK$ eigenvalues of $\mathbf M_K$, including the $L_K$ positive eigenvalues and the zero eigenvalues, by $\alpha_{1,K},\ldots,\alpha_{m_uK,K}$. Since both the number and the values of these gains depend on $K$, we characterize their long-horizon behavior through their distribution. Specifically, we show that the distribution of the finite-horizon gains approaches that of the spectral gains $\lambda_1(\omega),\ldots,\lambda_{m_u}(\omega)$ as $K$ increases. This distributional convergence is illustrated in Fig.~4(a).

For a frequency-domain matrix function $\mathbf F(\mathrm e^{\mathrm j\omega})$ of $\omega$, let
\begin{equation}
\mathbf F_l
=
\frac{1}{2\pi}
\int_{-\pi}^{\pi}
\mathbf F(\mathrm e^{\mathrm j\omega})
\mathrm e^{\mathrm j\omega l}
\,\mathrm d\omega
\end{equation}
denote its Fourier coefficients. The corresponding $K$-block Toeplitz section \cite{delsarte1991tridiagonal} is
\begin{equation}
\mathcal T_K(\mathbf F)
=
[\mathbf F_{r-s}]_{r,s=0}^{K-1}.
\end{equation}
Under the stationary model,
$\mathbf K_{Z,K}=\mathcal T_K(\mathbf S_Z)$ and
$\mathbf Q_K=\mathcal T_K(\mathbf Q)$.

The finite-section regularity condition in Theorem~2 requires the finite-horizon matrix $\mathbf M_K$ to have the same asymptotic spectral characterization as the Toeplitz matrix generated by $\mathbf L(\mathrm e^{\mathrm j\omega})$. Specifically,
\begin{align}
\sup_K
\left(
\|\mathbf M_K\|_2+
\|\mathcal T_K(\mathbf L)\|_2
\right)
&<\infty,
\label{eq:app_norm_bound}\\
\lim_{K\to\infty}
\frac{
\left\|\mathbf M_K-\mathcal T_K(\mathbf L)\right\|^2_{\mathrm F}
}{
K
}
=0.
\label{eq:app_finite_section}
\end{align}
In the stable LTI setting of Section~IV, this condition follows from standard finite-section results when the maneuver-to-observation impulse response is absolutely summable and $\mathbf S_Z$ and $\mathbf Q$ are continuous and uniformly positive definite \cite{gutierrez2008}.

Let $\beta_{1,K},\ldots,\beta_{m_uK,K}$ denote the eigenvalues of $\mathcal T_K(\mathbf L)$, ordered consistently with those of $\mathbf M_K$. Since both matrices are Hermitian, the Hoffman--Wielandt inequality yields \cite{hoffman1953variation}
\begin{equation}
\lim_{K\to\infty} \frac{1}{K} \sum_{i=1}^{m_uK} |\alpha_{i,K}-\beta_{i,K}|^2=0.
\label{eq:app_eigenvalue_difference}
\end{equation}
The uniform norm bound places both sets of eigenvalues in a common compact interval $[0,\Lambda]$. Any continuous function $f$ is uniformly continuous on this interval. Together with \eqref{eq:app_eigenvalue_difference}, this implies
\begin{equation}
\lim_{K\to\infty} \frac{1}{K} \sum_{i=1}^{m_uK} \left[ f(\alpha_{i,K})-f(\beta_{i,K}) \right]=0.
\label{eq:app_function_difference}
\end{equation}
Applying the block Szeg\H{o} eigenvalue-distribution theorem \cite[Th.~3]{gazzah2001asymptotic} gives
\begin{equation}
 \lim_{K\to\infty} \frac{1}{K} \sum_{i=1}^{m_uK} f(\alpha_{i,K}) = \frac{1}{2\pi} \int_{-\pi}^{\pi} \sum_{\iota=1}^{m_u} f\!\left(\lambda_\iota(\omega)\right) \,\mathrm d\omega.
\label{eq:app_gain_distribution}
\end{equation}

Equation~\eqref{eq:app_gain_distribution} shows that, for any continuous function \(f\), the normalized sum of \(f\) evaluated at the finite-horizon gains converges to the corresponding frequency-domain average over the spectral gains. Equivalently, after normalization by $m_u$, the empirical distribution of the finite-horizon gains approaches the distribution generated by $\{\lambda_\iota(\omega)\}_{\iota=1}^{m_u}$ as the horizon grows.

\subsection{Convergence of the Water Level}

We next show that the finite-horizon water-filling allocation converges to the spectral allocation in (29). For a horizon of $K$ time steps, water-filling over all $m_uK$ gains gives
\begin{equation}
p_{i,K}^{\star}
=
\left(
\mu_K-\frac{1}{\alpha_{i,K}}
\right)^+,
\qquad
\frac{1}{K}
\sum_{i=1}^{m_uK}
p_{i,K}^{\star}
=
P_s,
\label{eq:app_finite_wf}
\end{equation}
where $p_{i,K}^{\star}=0$ when $\alpha_{i,K}=0$.

Since the spectral gains are positive on a set of frequencies with nonzero bandwidth, there exist constants $a_0>0$ and $c>0$ such that, for sufficiently large $K$, at least $cK$ finite-horizon gains satisfy $\alpha_{i,K}\ge a_0$. It follows that
\begin{align}
P_s
\ge
c
\left(
\mu_K-\frac{1}{a_0}
\right)^+,
\end{align}
which establishes that the sequence $\{\mu_K\}$ is bounded.

To identify the possible subsequential limits of $\{\mu_K\}$, let $\{\mu_{K_q}\}$ be any convergent subsequence and write
\begin{equation}
\lim_{q\to\infty}\mu_{K_q}=\bar\mu.
\end{equation}
Define
\begin{equation}
h_\mu(a)
=
\begin{cases}
\left(\mu-\dfrac{1}{a}\right)^+,
& a>0,\\[1mm]
0,
& a=0.
\end{cases}
\end{equation}
For every finite $\mu$, $h_\mu(a)$ is continuous on $[0,\Lambda]$. Moreover,
\begin{equation}
\sup_{a\in[0,\Lambda]}
|h_{\mu_{K_q}}(a)-h_{\bar\mu}(a)|
\le
|\mu_{K_q}-\bar\mu|.
\label{eq:app_uniform_h}
\end{equation}
Thus, replacing $h_{\mu_{K_q}}$ with the fixed function $h_{\bar\mu}$ in the normalized resource sum introduces an error that vanishes in the limit.

Using the finite-horizon resource equality \eqref{eq:app_finite_wf} and applying \eqref{eq:app_gain_distribution} to the fixed continuous function $h_{\bar\mu}$ yield
\begin{equation}
G(\bar\mu)
\triangleq
\frac{1}{2\pi}
\int_{-\pi}^{\pi}
\sum_{\iota=1}^{m_u}
h_{\bar\mu}\!\left(\lambda_\iota(\omega)\right)
\,\mathrm d\omega
=
P_s.
\label{eq:app_limit_resource}
\end{equation}

The function $G(\bar\mu$ is continuous and nondecreasing in $\bar\mu$. Therefore, for every $P_s>0$, there is a unique water level $\mu$ satisfying $G(\mu)=P_s$. Since every convergent subsequence of the bounded sequence $\{\mu_K\}$ has the same limit $\mu$, we obtain
\begin{equation}
\lim_{K\to\infty}\mu_K=\mu.
\label{eq:app_mu_convergence}
\end{equation}
This water level $\mu$ defines the spectral water-filling allocation
\begin{equation}
p_\iota^\star(\omega)
=
\left(
\mu-\frac{1}{\lambda_\iota(\omega)}
\right)^+,
\label{eq:app_limit_wf}
\end{equation}
with $p_\iota^\star(\omega)=0$ whenever
$\lambda_\iota(\omega)=0$.

\subsection{Convergence of the Mobility Information}

We now evaluate the information associated with the converging finite-horizon allocation. In terms of all $m_uK$ gains, the normalized finite-horizon capacity is
\begin{equation}
\frac{C_M^{\mathrm G}(K\Delta t)}{K}
=
\frac{1}{2K}
\sum_{i=1}^{m_uK}
\log_2
\left(
1+\alpha_{i,K}p_{i,K}^{\star}
\right).
\label{eq:app_finite_capacity}
\end{equation}

Define
\begin{equation}
g_\mu(a)
=
\log_2\!\left(
1+a h_\mu(a)
\right).
\end{equation}
On the bounded interval $[0,\Lambda]$,
\eqref{eq:app_mu_convergence} implies
\begin{equation}
\sup_{a\in[0,\Lambda]}
|g_{\mu_K}(a)-g_\mu(a)|
\le
\frac{\Lambda}{\ln 2}
|\mu_K-\mu|
\longrightarrow 0.
\label{eq:app_uniform_g}
\end{equation}
We can therefore replace the $K$-dependent function with its limit and apply \eqref{eq:app_gain_distribution}. Equation~\eqref{eq:app_finite_capacity} then yields
\begin{align}
\lim_{K\to\infty}
\frac{C_M^{\mathrm G}(K\Delta t)}{K}
&=
\frac{1}{4\pi}
\int_{-\pi}^{\pi}
\sum_{\iota=1}^{m_u}
\log_2\!\left(
1+
\lambda_\iota(\omega)
p_\iota^\star(\omega)
\right)
\,\mathrm d\omega,\nonumber
\label{eq:app_rate_limit}
\end{align}
which gives $C_\infty(P_s)$ in (27). By \eqref{eq:app_limit_resource}, the allocation in \eqref{eq:app_limit_wf} satisfies the KKT conditions of the spectral optimization problem in (27) and (28). The concavity of this problem establishes global optimality. This proves (27) and (29) when the spectral mobility gains are positive on a set of frequencies with nonzero bandwidth.

It remains to consider the case where $\lambda_\iota(\omega)=0$ almost everywhere for every $\iota$.
For any $\delta>0$, let $n_K(\delta)$ denote the number of finite-horizon gains exceeding $\delta$. Applying \eqref{eq:app_gain_distribution} with $f(a)=a$ gives
\begin{align}
\lim_{K\to\infty}
\frac{1}{K}
\sum_{i=1}^{m_uK}\alpha_{i,K}=0,
\end{align}
and hence $\lim_{K\to\infty}n_K(\delta)/K=0$ for every fixed $\delta>0$. Separating the contributions from gains below and above $\delta$ yields
\begin{equation}
\frac{C_M^{\mathrm G}(K\Delta t)}{K}
\le
\frac{\delta P_s}{2\ln2}
+
\frac{n_K(\delta)}{2K}
\log_2\!\left(
1+\frac{\Lambda P_s K}{n_K(\delta)}
\right).
\end{equation}
Since
$\lim_{x\to 0^+}x\log(1+c/x)=0$,
\[
\limsup_{K\to\infty}
\frac{C_M^{\mathrm G}(K\Delta t)}{K}
\le
\frac{\delta P_s}{2\ln2}.
\]
Letting $\delta\to 0^+$ and using the nonnegativity of mutual information yield
\begin{align}
    \lim_{K\to\infty}
\frac{C_M^{\mathrm G}(K\Delta t)}{K}=0.
\end{align}
Thus, $C_\infty(P_s)=0$ when all spectral mobility gains equal $0$ almost everywhere.

Finally, the rate relation
\begin{align}
R_M
= \lim_{K\to\infty} \frac{C_M^{\mathrm G}(K\Delta t)}{K\Delta t}=
\frac{C_\infty(P_s)}{\Delta t}
\end{align}
completes the proof of Theorem~2.

\bibliographystyle{IEEEtran}
\bibliography{references}